\documentclass[journal]{IEEEtran}

\ifCLASSINFOpdf
\else
\fi

\usepackage{graphicx}
\usepackage{multirow}
\usepackage{amsmath}
\usepackage{booktabs}
\usepackage{subfig}
\usepackage[export]{adjustbox}
\usepackage{placeins}
\usepackage{blindtext}
\usepackage{cite}
\usepackage{hyperref}
\usepackage{verbatim}
\usepackage{lipsum}
\usepackage{todonotes}
\usepackage{tabularx}
\usepackage{makecell}
\usepackage{stfloats}
\usepackage{overpic}
\definecolor{midnightblue}{rgb}{0.1, 0.1, 0.44}

\definecolor{mdgreen}{rgb}{0,0.7,0}

\definecolor{mdblue}{rgb}{0.2,0.3,0.95}

\begin{document}
\title{Burst Photography for Learning to Enhance Extremely Dark Images}

\author{Ahmet~Serdar~Karadeniz,
        Erkut~Erdem,
        Aykut~Erdem%
}

\maketitle

\IEEEpeerreviewmaketitle

\begin{abstract}
Capturing images under extremely low-light conditions poses significant challenges for the standard camera pipeline. Images become too dark and too noisy, which makes traditional enhancement techniques almost impossible to apply. Recently, learning-based approaches have shown very promising results for this task since they have substantially more expressive capabilities to allow for improved quality. Motivated by these studies, in this paper, we aim to leverage burst photography to boost the performance and obtain much sharper and more accurate RGB images from extremely dark raw images. The backbone of our proposed framework is a novel coarse-to-fine network architecture that generates high-quality outputs progressively. The coarse network predicts a low-resolution, denoised raw image, which is then fed to the fine network to recover fine-scale details and realistic textures. To further reduce the noise level and improve the color accuracy, we extend this network to a permutation invariant structure so that it takes a burst of low-light images as input and merges information from multiple images at the feature-level. Our experiments demonstrate that our approach leads to perceptually more pleasing results than the state-of-the-art methods by producing more detailed and considerably higher quality images. 

\end{abstract}


\section{Introduction}

Capturing images in low-light conditions is a challenging task -- the main difficulty being that the level of the signal measured by the camera sensors is generally much lower than the noise in the measurements~\cite{Liba2019}. The fundamental factors causing the noise are the variations in the number of photons entering the camera lens and the sensor-based measurement errors occurred when reading the signal~\cite{Hasinoff2014, Brooks2019}. In addition, noise present in a low-light image also affects various image characteristics such as fine-scale structures and color balance, further degrading the image quality. 

\begin{figure}[!t]
    \subfloat{
    \includegraphics[width=\linewidth]{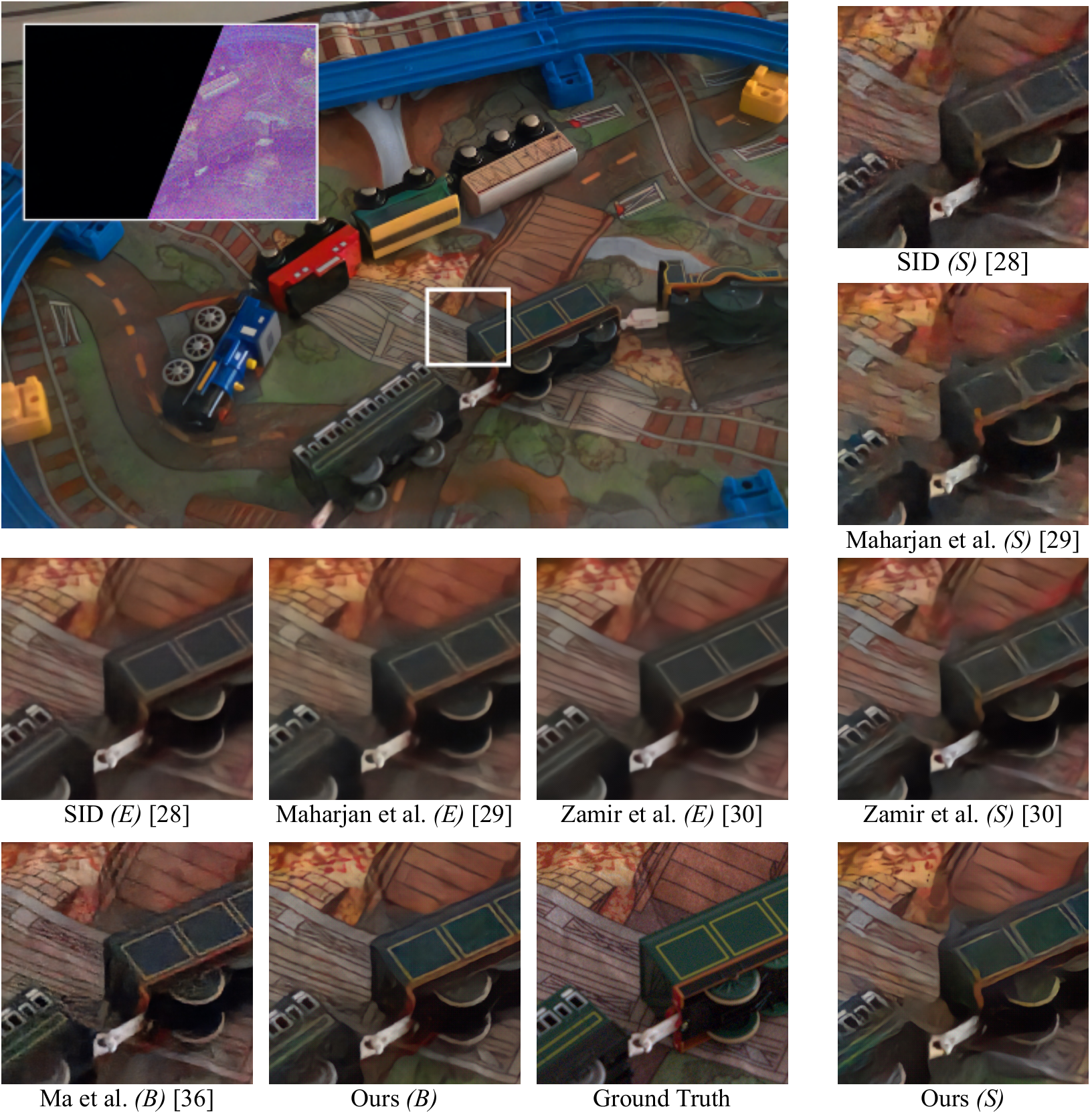}}
    \caption{A sample result obtained with our proposed burst-based extremely low-light image enhancement method. The standard camera output and its scaled version are shown at the top left corner. For comparison, the zoomed-in details from the outputs produced by the existing approaches are given in the subfigures. The results of the single image enhancement models, denoted with \textit{(S)}, are shown on the right. The results of the multiple image enhancement methods are presented at the bottom, with \textit{(B)} denoting the burst and \textit{(E)} indicating the ensemble models. Our single image model recovers finer-scale details much better than its state-of-the-art counterparts. Moreover, our burst model gives perceptually the most satisfactory result, compared to all the other methods.}
    \label{fig:teaser}
\end{figure}

Direct approaches for capturing bright photos in low light conditions include widening the aperture of the camera lens, lengthening the exposure time, or using camera flash~\cite{Liba2019, hasinoff2016burst}. These methods, however, do not solve the problem completely as each of these hacks has its own drawbacks. Opening the aperture is limited by the hardware constraints, and when the camera flash is used, the objects closer to the camera are brightened more than the objects or the scene elements that are far away \cite{petschnigg2004digital}. Images captured with long exposure times might have unwanted image blur due to camera shake or object movements in the scene~\cite{Sugimura2015}. Hence, in the literature, there has been a wide range of studies which try to improve the quality of low-light images, ranging from traditional denoising and enhancement methods to learning-based approaches.

Image denoising is one of the classical problems in image processing, where the aim is to restore a clean image from a noisy image. Several methods have been proposed over the years to denoise images~\cite{buades2005non,dabov2007image,talebi2013global,chang2000adaptive, elad2006image, rudin1992nonlinear, Jain2009,Xie2012,zhang2017beyond, zhang2018ffdnet,lehtinen2018noise2noise, krull2019noise2void,laine2019high}.
Most of these approaches rely on the images with Gaussian noise for developing a denoising model. Recently, deep learning-based methods that can deal with real image noise have been proposed~\cite{Brooks2019, guo2019toward}. However, these approaches are not specialized to extremely low-light images which are harder to restore than a standard noisy image. Image enhancement is another active field of research, which has seen tremendous progress in the past few years with deep learning~\cite{Lore2017,Tao2017, Lv2018,wang2019underexposed, wei2018deep,jiang2019enlightengan,guo2020zero}. Usually, these methods work with low dynamic range (LDR) input images and hence, their performance is also limited due to the errors accumulated in the camera processing pipeline. When compared to LDR images, raw images straight from the camera are more suitable to use for enhancing extremely low-light images since they contain more information and are processed minimally.

In the context of enhancing extremely dark images, See-in-the-Dark (SID)~\cite{Chen2018} is the first learning-based attempt to replace the standard camera pipeline, training a convolutional neural network (CNN) model to produce an enhanced RGB image from a single raw low-light image. For this purpose, the authors collected a dataset of short-exposure, dark raw photos and their corresponding long-exposure references. Their method is further improved by Maharjan et al.~\cite{Maharjan2019a} and Zamir et al.~\cite{Zamir2019a} with some changes in the CNN architecture and the objective functions utilized in training. In a similar fashion, in our study, we develop a new multi-scale architecture for single image enhancement and use a different objective by combining contextual and pixel-wise losses. While the previous methods obtain an RGB image from a single dark raw image, we further explore whether the results can be improved by integrating multiple observations regarding the scene.

Bracketing is a well-known technique in photography that relies on rapidly taking several shots of the same scene. These shots usually differ from each other in terms of some camera settings, e.g. exposure, which capture characteristics of the scene differently, and thus they can be used for applications like constructing a high dynamic range (HDR) image. A technique similar to exposure bracketing is shooting each frame in the burst sequence with a constant exposure~\cite{hasinoff2016burst}. To our interest, when shot with a constant short exposure under low-light, these images represent different dark, noisy realizations of the same scene. Naturally, they provide us multiple observations about the scene when compared to a single dark image. While simply averaging these images reduces noise, results are not always satisfactory. For this reason, different techniques are introduced to merge the temporal pixels in the burst sequence
\cite{Liba2019, hasinoff2016burst, buades2009note, joshi2010seeing, liu2014fast, Mildenhall2018, Godard2018, Ma2020}. Among these approaches, \cite{Mildenhall2018, Godard2018, Ma2020} use learning-based methods to process burst images. In these studies, burst images are fed to a CNN either by concatenating through channels or in a recurrent fashion.
In our case, we propose a radically different approach and show that processing these burst images in a permutation invariant manner is a simple yet more effective approach. The order of burst images does not affect the output, and accordingly a more accurate output can be obtained. In Fig.~\ref{fig:teaser}, we present the results of the aforementioned extremely low-light image enhancement models along with our results. The multiple image enhancement models, which either employ burst imagery or integrate ensemble of enhanced images, give superior results than their single image counterparts, yet they still suffer from artifacts such as over-smoothing, and fail to recover fine-scale details in the image. Despite the remarkable progress of previous studies~\cite{Chen2018,Maharjan2019a,Zamir2019a,Ma2020}, this example image demonstrates that there is still large room for improvement, regarding various issues such as unwanted blur, noise and color inaccuracies in the end results -- especially for the input images which are extremely dark.

In a nutshell, to alleviate these shortcomings, in this study, we propose a learning-based framework that takes a burst of extremely low-light raw images of a scene as input and generates an enhanced RGB image. In particular, we develop a coarse-to-fine network architecture which allows for simultaneous processing of a burst of dark raw images as input to obtain a high quality RGB image.

Our main contributions are summarized as follows:
\begin{itemize}
  \item We introduce a multi-scale deep architecture for image enhancement under extremely dark lighting conditions, which consists of a coarse-scale network and a fine-scale network. 
  \item We further extend our coarse-to-fine architecture to design a novel permutation invariant CNN model that predicts an enhanced RGB image by integrating features from a burst of images of a dark scene.
  \item Our experiments demonstrate that our approach outputs RGB images with less noise and sharper edge details than those of the state-of-the-art methods. These are validated quantitatively based on several quality measures in both single-frame and burst settings.
  \item We also show that the proposed burst model can be applied to videos taken at dark environments involving dynamic objects by additionally incorporating a motion compensation module based on optical flows predicted from the outputs at the coarse-scale network.
\end{itemize}
Our models are publicly available at the project webpage: {\color {midnightblue}\href{https://hucvl.github.io/dark-burst-photography/}{https://hucvl.github.io/dark-burst-photography/}}.

\section{Related Work}
Low-light images show different characteristics due to the lighting conditions of the environments, and the noise and/or motion blur they contain. In general, the approaches for low-light image processing can be divided into two groups, with respect to the darkness levels of the input images: (i) low-light image enhancement, and (ii) extremely low-light image enhancement. Generic low-light image enhancement methods refer to the approaches that restore the perceptual quality of images taken under poor illumination conditions, which suffer from low visibility. Enhancement models for extremely low-light images, on the other hand, deal with images captured under more severe conditions, which cannot be handled by the first group of works. In particular, the darkness of an image is directly related to the illuminance of a scene, which is measured in terms of lumens per meter squared (lux). In this sense, extremely low light images denote short exposure images (usually between 1/30 and 1/10 sec exposure) that are taken in 0.2-5 lux outdoor or 0.03-0.3 lux indoor scenes.

In this study, we explore the use of burst photography for enhancing extremely dark images. Since extremely low-light images contain severe noise, our work is also related to generic image denoising and burst photography. Hence, in this section, we provide a brief review of image denoising, low-light image enhancement, extremely low-light image enhancement and burst photography methods proposed in recent years.

\begin{figure}[!t]
\begin{tabular}{c@{$\;$}c}
\includegraphics[width=0.48\linewidth]{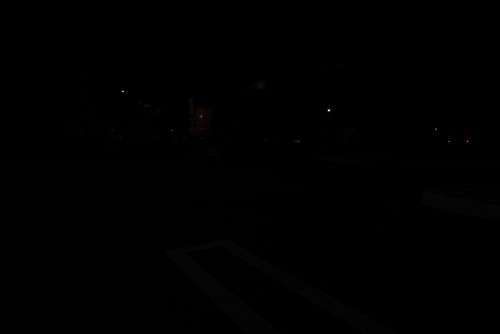} &
\includegraphics[width=0.48\linewidth]{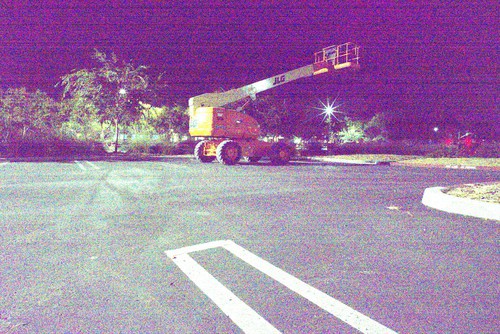} \vspace{-0.1cm}\\
{\footnotesize (a) Dark} & {\footnotesize (b) Traditional} \vspace{0.1cm}\\
\includegraphics[width=0.48\linewidth]{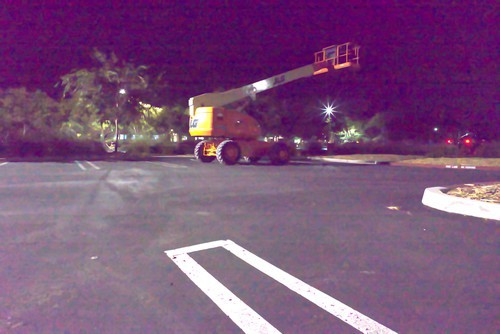} &
\includegraphics[width=0.48\linewidth]{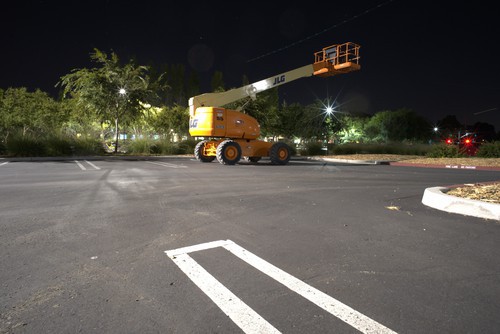} \vspace{-0.1cm}\\
{\footnotesize (c) Traditional + BM3D denoising} &
{\footnotesize (d) Long exposure}
\end{tabular}
\caption{For an extremely dark image displayed in (a), the traditional camera pipeline produces a highly noisy image with severe color degradation, as shown in (b). Moreover, as demonstrated in (c), the state-of-the-art denoising methods cannot handle these challenges and give unsatisfactory results. Extremely low-light image enhancement methods, on the other hand, aim for generating an output close to a long-exposure image, like the one given in (d).}
\label{fig:darkness}
\end{figure}

\subsection{Image Denoising}
Image denoising is a fundamental problem in computer vision that deals with removing noise from an image~\cite{Gu2019,Chatterjee2010}. Traditionally, methods that exploit the non-local self-similarity prior \cite{buades2005non,dabov2007image,talebi2013global}, sparsity \cite{chang2000adaptive, elad2006image} and image gradients \cite{rudin1992nonlinear} have been widely used for image denoising. Recently, various deep learning approaches have been proposed for both non-blind Gaussian denoising~\cite{Jain2009,Xie2012} and blind Gaussian denoising \cite{zhang2017beyond, zhang2018ffdnet}, which involve training denoising models under known and unknown noise levels, respectively. Lately, researchers proposed unsupervised deep denoising models \cite{lehtinen2018noise2noise, krull2019noise2void, laine2019high} that do not use any clean ground truth data during training. Although most of these existing denoising models focus on additive white Gaussian noise, this noise model falls short when the real-life images are considered. Hence, the recent trend in image denoising is to develop models that are trained with real-world noisy data \cite{guo2019toward, Brooks2019} and that can generalize much better than the models which consider additive white Gaussian noise. While these aforementioned recent methods give fairly good results most of the time, they are not well-suited to extremely dark images as they suffer from severe noise and color degradation, as shown in Fig.~\ref{fig:darkness}.

\begin{figure*}[!t]
\centering
\subfloat{
\includegraphics[width=\linewidth]{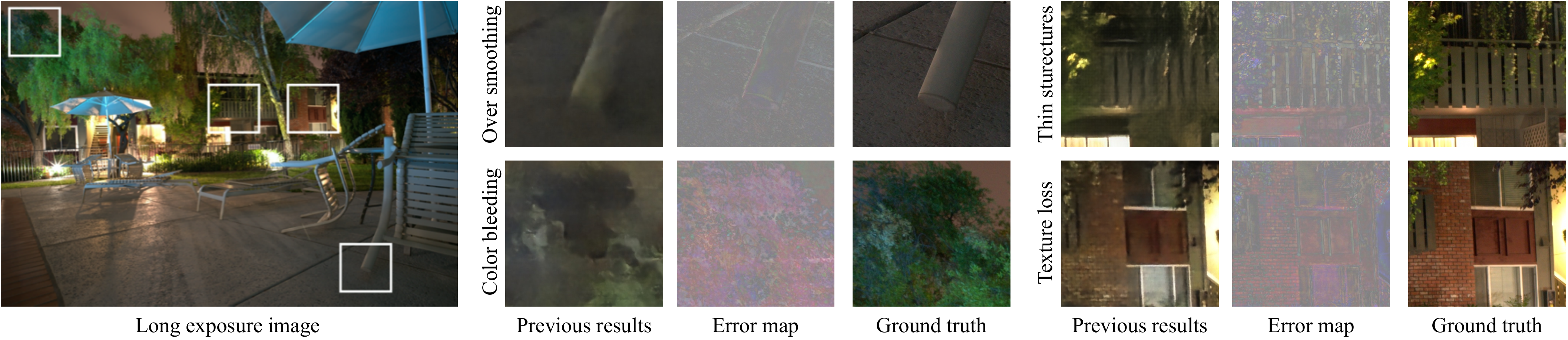}}
\caption{Common failure cases for the state-of-the-art extremely low-light image enhancement methods. Subfigures show some cropped images from the results of the existing models together with the corresponding error and the ground truth images, demonstrating that these models suffer from over-smoothing and color bleeding artifacts and fail to properly recover thin structures and textured regions.}
\label{fig:artifacts}
\end{figure*}

\subsection{Low-Light Image Enhancement}
Generic approaches that can be used for low-light image enhancement can be divided into three groups: (i) traditional contrast enhancement methods, (ii) techniques based on Retinex-theory, and (iii) learning-based approaches. Most well-known methods for contrast enhancement include histogram equalization based approaches that apply transformations to image histograms ~\cite{Hummel1977, Zuiderveld1994, Ibrahim2007, arici2009histogram}.
Motivated by human color perception, Retinex-theory based approaches decompose the images into illumination and reflectance components, and take into account these components while enhancing the images~\cite{Land1977, Ng2011, Fu2016a, Guo2017, jobson1997multiscale}.
On the other hand, learning-based methods mostly include discriminative methods based on sparse autoencoders \cite{Lore2017} and CNNs that either directly estimate an enhanced image \cite{Tao2017, Lv2018} or extract an illumination map \cite{wang2019underexposed, wei2018deep}. Recently, researchers suggested some unsupervised models which employ adversarial losses for enhancement \cite{jiang2019enlightengan} or CNNs for illumination curve estimation~\cite{guo2020zero}.

These low-light image enhancement methods provide good results under certain conditions. However, they fail to deal with the full extent of the challenges in imaging under extremely dark conditions. They mainly accept LDR images generated by the standard camera pipeline. Transforming raw images to LDR images introduces some information loss in the measurements which complicates the enhancement process. Hence, these low-light image enhancement models are favorable only when the input images are partly dark and do not exhibit serious color degradation and severe noise. 

\subsection{Extremely Low-Light Image Enhancement}
As discussed in the introduction, enhancing extremely dark images was introduced as a challenging image enhancement task by Chen et al. in~\cite{Chen2018}, and the See-in-the-Dark (SID) model proposed therein is the first model that specifically aims for solving this task. This approach processes a raw image captured under very poor illumination condition with a \mbox{U-Net}~\cite{RFB15a} like architecture. Training of the model is carried out on a dataset of paired short and long-exposure images by taking into account a pixel-wise ($L_\text{1}$) loss. 

Very recently, there have been attempts to further improve the performance of SID. Maharjan et al.~\cite{Maharjan2019a} have proposed to use residual learning to boost the final image quality. Zamir et al.~\cite{Zamir2019a} have used a hybrid loss function which is a combination of pixel-wise and multi-scale structural similarity (MS-SSIM) losses and a perceptual loss~\cite{Dosovitskiy2016,JohnsonECCV2016}, which is defined by the absolute difference of the features extracted by a deep network. Interestingly, in~\cite{Ma2020}, Ma et~al. have developed an enhancement model for extremely low-light images, which employs recurrent convolutional neural networks to obtain a high quality result from a burst of input images. Although these works demonstrate progress in enhancing extremely low-light images, they do not completely solve the challenges of the dark scenes. As presented in Fig.~\ref{fig:artifacts}, over-smoothing, color bleeding, recovery of thin structures or textures remain as the main difficulties of enhancing dark images.

As will be discussed in the next section, different from the aforementioned methods, we alternatively propose a multi-scale approach which uses a novel coarse-to-fine architecture that better handles the extremely low-light images by giving much sharper and more vivid colors. In addition, we use a combination of the $L_1$ pixel loss and the recently proposed contextual loss function which maintains the image statistics better~\cite{mechrez2018maintaining}. Moreover, for our burst model, we employ a set-based permutation invariant architecture that jointly processes low-light input images in an orderless manner, giving perceptually plausible and high quality results.

There are also some recent efforts to extend the aforementioned image enhancement models to videos by additionally taking into account temporal consistencies. For instance, Chen et al.~\cite{chen2019seeing} extended their SID model to videos by training a Siamese network on static raw videos. Similarly, Jiang and Zheng~\cite{jiang2019motion} proposed a U-Net like architecture containing 3D convolution layers for the same purpose. In~\cite{MaSiggraph2020}, Ma~et~al. presented a new computational
photography technique with single-photon cameras that   allows for generating high-quality images under fast motion and extremely low light conditions.

\subsection{Burst Photography}
Burst photography refers to the process of capturing a sequence of images each spaced a few milliseconds apart and subsequently integrating them to obtain a higher-quality image. For instance, the most intuitive way to produce a noise-free image is to capture a burst of images and apply simple averaging. Yet, this strategy gives unsatisfactory results in practice due to moving objects and/or a moving camera. Hence, a variety of more complicated methods were introduced to combine the information from multiple images in a more effective manner. Buades et al. proposed to apply standard averaging only for the aligned pixels and utilize the state-of-the art denoising methods for the remaining pixels~\cite{buades2009note}. Joshi et al. developed a method that weights the pixels with respect to their sharpness levels by using Laplacian convolution \cite{joshi2010seeing} and accordingly utilizes these weights in obtaining higher quality images. Liu et al. proposed to fuse the consistent pixels with an optimal linear estimator~\cite{liu2014fast}. Moreover, some researchers suggested to employ the information encoded in the frequency-domain for temporal fusion~\cite{hasinoff2016burst, Liba2019, delbracio2015hand}. Recently, more sophisticated approaches have been proposed for denoising such as Kernel Prediction Networks \cite{Mildenhall2018}, Recurrent Fully Convolutional Networks \cite{Godard2018}, and Permutation Invariant Networks \cite{Aittala2018}, which process a burst of noisy and blurred images through deep CNN architectures.

\begin{figure*}[!t]
\centering
\subfloat[Coarse-to-fine network]{
\includegraphics[width=0.49\linewidth]{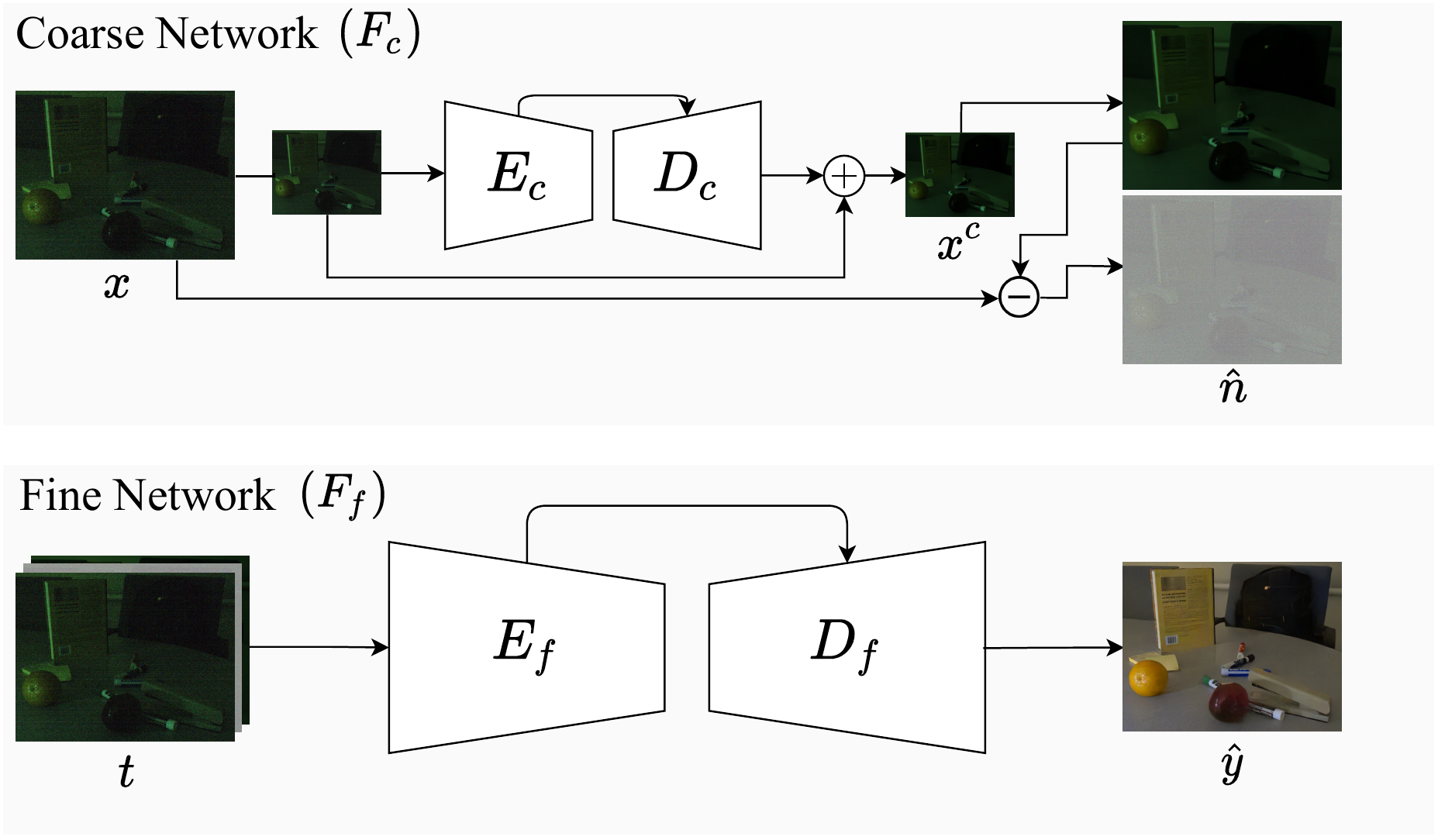}}
\subfloat[Set-based network]{
\includegraphics[width=0.49\linewidth]{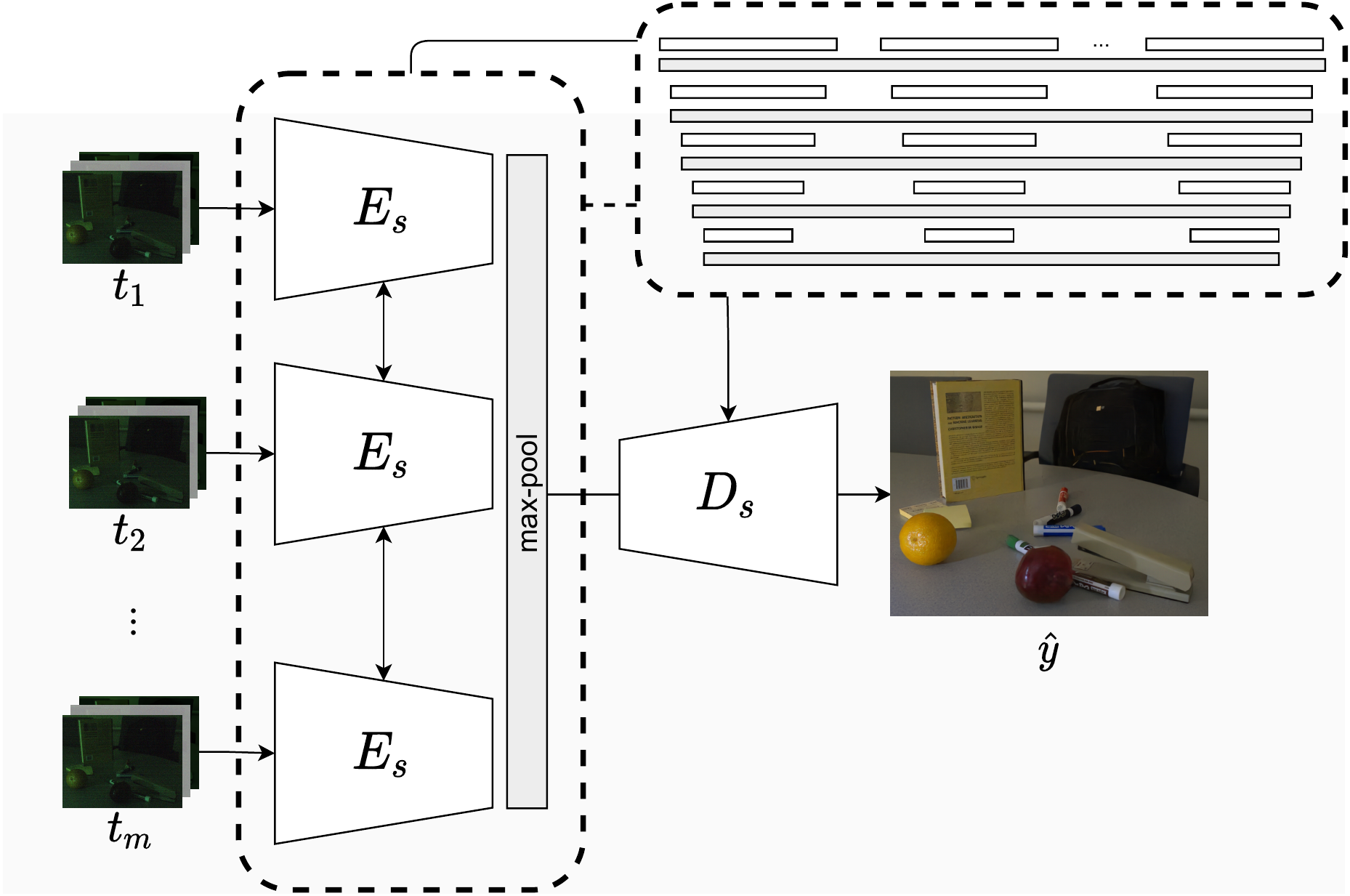}}

\caption{Network architectures of the proposed (a) single-frame coarse-to-fine model, and (b) set-based burst model.}
\label{fig:proposed}
\end{figure*}

These aforementioned models do not cope with the challenges of extremely dark images -- with the exception of Liba et al.~\cite{Liba2019} and Hasinoff et al. \cite{hasinoff2016burst} where the authors rely on hand-crafted strategies. As mentioned before, the only work that focuses on learning-based burst imagery in the extremely low-light conditions is the work by Ma et al.~\cite{Ma2020}. In this work, the authors utilized a recurrent convolutional neural network architecture, similar to the one in~\cite{Godard2018}, to enhance a burst of raw low-light images. In our work, specifically motivated by these recent burst photography approaches, we develop a set-based permutation invariant CNN architecture that can be used to obtain a high quality image from a burst of extremely dark images. In particular, as compared to the recurrent model in~\cite{Ma2020} which processes each frame sequentially, our network jointly processes the burst frames in an orderless manner.

\section{Our Approach}
Table~\ref{tab:notations} summarizes the notations used in the paper. Our aim is to learn a mapping from the domain of raw low-light images to the domain of long-exposure RGB images. To achieve this, we first propose a single-frame coarse-to-fine model and then extend it to a set-based formulation to process a burst of images. The details of our networks are illustrated in Fig.~\ref{fig:proposed}. 

\renewcommand{\arraystretch}{1.15}
\begin{table}[!t]
\begin{center}
\begin{tabular}{p{2.7cm}@{}p{6cm}}
\toprule
$x_1, x_2, \dots, x_m$ & Burst of raw low-light input images \\
$y$, $\hat{y}$ & Reference and predicted long-exposure RGB images\\
$F_{c}(\cdot), F_{f}(\cdot), F_\text{s}(\cdot)$ & Coarse, fine and set-based networks \\
$x_1^{c}, x_2^{c}, \dots, x_m^{c}$ & Raw, low-res outputs of the coarse network \\
$\hat{n}_1, \hat{n}_2, \dots, \hat{n}_m$ &
Noise approximations for $x_1, x_2, \dots, x_m$ \\
$t_1, t_2, \dots, t_m$ & Tensors containing raw inputs, upsampled coarse outputs and noise approximations \\
$R_d(\cdot), R_u(\cdot)$ & Downsampling and upsampling functions\\
\bottomrule
\end{tabular}
\end{center}
\caption{The notations used throughout the paper.}
\label{tab:notations}
\end{table}
\renewcommand{\arraystretch}{1}

\subsection{Coarse-to-fine Model}
To recover fine-grained details from dark images, we propose to employ a two-step coarse-to-fine training procedure. The proposed architecture consists of a coarse network and a subsequent fine network. The purpose of the coarse network is to obtain a coarse enhancement result and an approximate noise map in lower resolutions, which can be considered as extra guidance data for the subsequent fine network. That is, the fine network processes the dark input image under the guidance of these two to obtain a refined output. Considering low-resolution inputs and outputs in our coarse network mainly speeds up the processing time and decreases the memory footprint of our overall framework.
Similar strategies have been proven very effective in various other tasks such as deblurring \cite{Nah2017} and image synthesis \cite{wang2018high}. Different than those approaches, our coarse network outputs a raw (denoised and enhanced) image. Predicting the coarse outputs in the raw domain also allows us to compute an approximate noise map from the input. Similar to some recent denoising methods~\cite{Brooks2019, Mildenhall2018}, we use this noise map as an additional input for the second stage of our framework. This introduces certain inductive biases into the model architecture that allows for extracting more fine-grained features that reflect the structure of the underlying extremely dark input image better.

In our proposed framework, the raw low-light input image is first downsampled by a factor of two and then fed to our coarse network. The coarse network, which is illustrated in Fig.~\ref{fig:proposed}(a), is trained on downsampled data and produces denoised and enhanced outputs in low-resolution
\begin{equation}
    x^{c} = F_{c}(R_d(x)).     
\end{equation}

We utilize the output of the coarse network not just for guidance in assisting the fine network but also in approximating the noise by computing the difference between the upsampled coarse prediction and the raw low-light input, as:
\begin{equation}
    \hat{n} = x-R_u(x^c)
\end{equation}
The fine network takes the concatenation of the low-light raw input image, the output from the coarse network and the noise approximation as inputs and processes them to generate the final RGB output:
\begin{equation}
    \hat{y} = F_{f}(t), \quad t = (x, \hat{n}, R_u(x^c)) 
\end{equation}

Both our coarse and fine networks follow a U-Net like encoder-decoder architecture. In the encoder, they contain 10 convolution layers where the number of filters is doubled and the resolution is halved after every 2 convolution layers, with the initial number of filters is set to 32. In the decoder, they include deconvolution layers which are concatenated with earlier corresponding convolution layers through skip connections. Similar to the models in~\cite{lai2018learning, chen2019seeing}, there are 16 residual blocks between the encoder and decoder of the fine network. Additionally, squeeze-excitation layers~\cite{hu2018squeeze, Maharjan2019a} are added to the residual block before the summation with the identity branch. As shown in Fig.~\ref{fig:main}, the coarse network gives a fairly good enhancement result for a given extremely low-light image containing severe noise and color degradation. The fine network further improves the color accuracy and the details of the coarse network's result, producing a higher quality image.

\subsection{Set-Based Extension to Burst Images}
Recently, there have been some attempts to study the invariance and equivariance properties of neural networks \cite{Ravanbakhsh2017, Cohen2016, Gens2014}. Zaheer et al. provided a generic method to train neural networks that operate on sets via a simple parameter sharing scheme~\cite{zaheer2017deep}, which allows for information exchange with a commutative operation. Based on this idea, Aittala and Durand proposed a permutation invariant CNN architecture for burst image deblurring~\cite{Aittala2018}. In a similar vein, we also design a permutation invariant CNN model, but with a lower computational cost using multiple encoders and a single decoder.

\begin{figure*}[!thb]
\centering
\captionsetup[subfigure]{labelformat=empty,farskip=1pt,captionskip=1pt}

\setcounter{subfigure}{0}
\captionsetup[subfigure]{labelformat=empty}

\subfloat[Traditional]{
\includegraphics[width=0.23\linewidth]{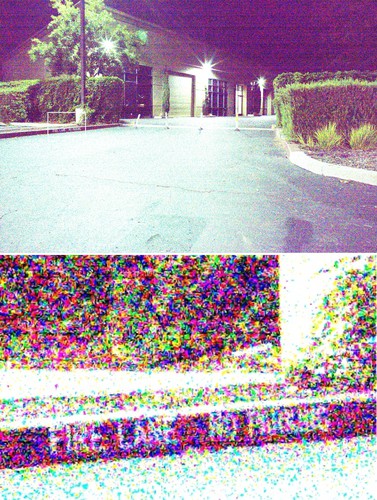}}
\subfloat[Coarse]{
\includegraphics[width=0.23\linewidth]{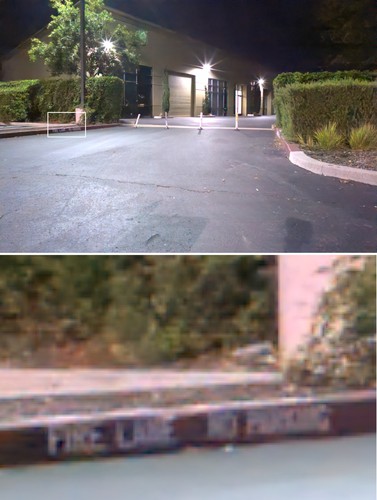}}
\subfloat[Fine]{
\includegraphics[width=0.23\linewidth]{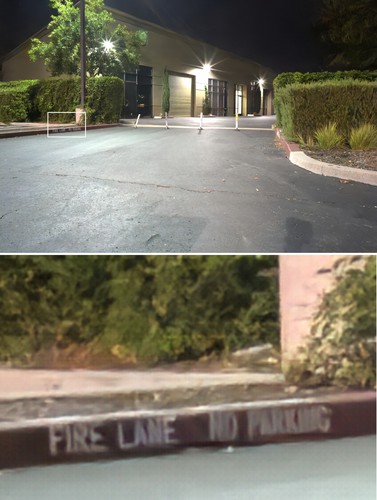}}
\subfloat[Burst]{
\includegraphics[width=0.23\linewidth]{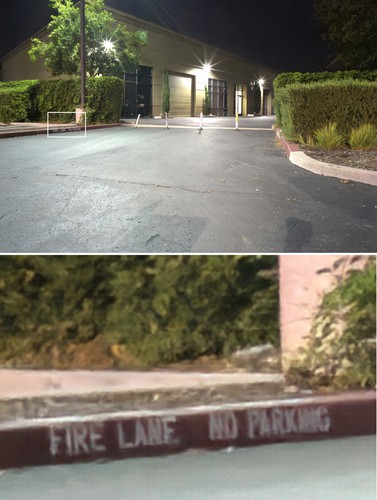}}
\caption{An example night photo captured with 0.1 sec exposure and its enhanced versions by the proposed coarse, fine and burst networks. As the cropped images demonstrate, the fine network enhances both the color and the details of the coarse result. The burst network produces even much sharper and perceptually more pleasing output.}
\label{fig:main}
\end{figure*}

We extend our coarse-to-fine model to a novel permutation invariant CNN architecture which takes multiple images of the scene as input and predicts an enhanced image. In particular, first, low-resolution coarse outputs are obtained for each frame $x_i$ in the burst sequence, using our coarse network:
\begin{equation}
    x_i^c = F_c(R_d(x_i))
\end{equation}
In addition, we compute an approximate noise component $n_i$ for each frame, as
\begin{eqnarray}
    \hat{n}_i = x_i - R_u(x_i^c) \;.
\end{eqnarray}

Finally, our set-based network accepts a set of tensors $\{t_i\}$ as input, each instance $t_i = \left(x_i, \hat{n}_i, R_u(x^c_i)\right)$ corresponding to the concatenation of one of raw burst images $x_i$, its noise approximation $\hat{n}_i$ and the upsampled version of the coarse prediction  $R_u(x^c_i)$, and produces the final RGB output:
\begin{equation}
    \hat{y} = F_s\left(\{t_1, \dots, t_m\}\right) \;.
 \end{equation}
Here, $F_s$ represents our permutation invariant CNN, which has $m$ convolutional subnetworks which share parameters with each other and allow for information exchange between the features of burst frames. This is achieved by using a max-pooling over the set of burst features after each convolution layer in the encoder part of the network. Then, in the decoder part, instead of concatenating the deconvolution features with the corresponding earlier features, we concatenate them with the global max-pooled features computed in the encoder. Hence, without even changing the parameter size, we integrate the advantage of multiple observations to the network.

To obtain robustness to small motions, we apply max fusion between the features of burst frames after the second convolution block. As the features are downsampled, their alignment becomes much easier and the network benefits from the fusion of the higher-level features. To deal with large motions in the scene, however, we can utilize the outputs of our coarse network to estimate optical flows between consecutive frames. In our experiments, we employ the method in~\cite{teed2020raft} to obtain the optical flow maps, which are then used to selectively zero out the regions with large motion\footnote{Here, we extract optical flows after converting the raw coarse Bayer data to raw RGB by splitting it into distinct RGB channels with the green channel obtained by averaging the two green subpixels in each two-by-two pattern.}. Thus, inputs to the fine network are these processed burst frames which differ from each other in the regions with small motion, and their fusion is performed via the max-pooling layers inside the network.

As Fig.~\ref{fig:main} demonstrates, processing multiple dark images via the proposed burst network significantly improves the quality. Our burst model produces perceptually better and sharper results than our fine network and especially recovers the fine details and the texture much better. We analyze its ability to handle motion in more detail in our experimental analysis.

\subsection{Losses}
To train our networks, we tested combining a pixel-wise loss ($L_{\text{1}}$) with two alternative featurewise losses, namely perceptual loss ($L_{\text{P}}$)~\cite{Dosovitskiy2016,JohnsonECCV2016} and contextual loss ($L_{\text{CX}}$)~\cite{mechrez2018maintaining, mechrez2018contextual}.

\textbf{Pixel-wise Loss.}
As the pixel-wise loss, we use the $L_\text{1}$ loss between the network output and the ground truth long-exposure image, given as:
\begin{equation}
{L}_1(y, \hat{y}) = \left\|{y - \hat{y}}\right\|_1.
\end{equation}

\textbf{Perceptual Loss.}
To measure the distance at a more semantic level, we employ the commonly used perceptual loss~\cite{Dosovitskiy2016,JohnsonECCV2016}, which uses high-level features from a pre-trained VGG-19 network~\cite{simonyan2014very}, defined as:
\begin{equation}
    L_\text{P}(y, \hat{y}, l) = \left\|\phi^l(y) - \phi^l(\hat{y})\right\|_1
\end{equation}
where $\phi^l(\cdot)$ denotes the feature maps at the $l$-th layer of the network. 

\textbf{Contextual Loss}. As an alternative to the perceptual loss, we also consider the contextual loss proposed in~\cite{mechrez2018maintaining, mechrez2018contextual}, which is shown to better capture changes in fine scale details. %
Specially, it measures the statistical difference between the feature distributions $\phi^l(y)$ and $\phi^l(\hat{y})$ extracted from $y$ and $\hat{y}$, respectively, and is defined as:
\begin{equation}
    L_\text{CX}(y, \hat{y}, l) = -\log(\text{CX}(\phi^l(y), \phi^l(\hat{y})))
\end{equation}
where the statistical similarity $\text{CX}$ is estimated by an approximation of the KL-divergence, as follows.

Let $R=\{r_i\}$ and $S=\{s_j\}$ respectively represent the set of features extracted from a pair of images, with cardinality~$N$, and $d_{ij}$ be the cosine distance between the features $r_i$ and $s_j$. %
Then, $\text{CX}(R,S) = \frac{1}{N} \sum_j{\underset{i}{\max}\;} \text{CX}_{ij}$ where \mbox{$\text{CX}_{ij} = w_{ij}/\sum_k{w_{ik}}$}
and $w_{ij} = \exp{(({1 - \Tilde{d}_{ij}})/{h})}$, \mbox{$\Tilde{d}_{ij} = {d_{ij}}/
{({\min_k\;} d_{ik} + \epsilon})$}.

\begin{figure*}[!thb]
    \centering
\captionsetup[subfigure]{labelformat=empty,farskip=1pt,captionskip=1pt}
    
\subfloat{
\includegraphics[width=0.18\linewidth]{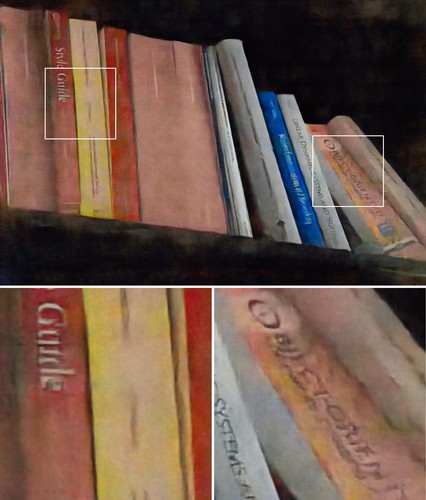}}
\subfloat{
\includegraphics[width=0.18\linewidth]{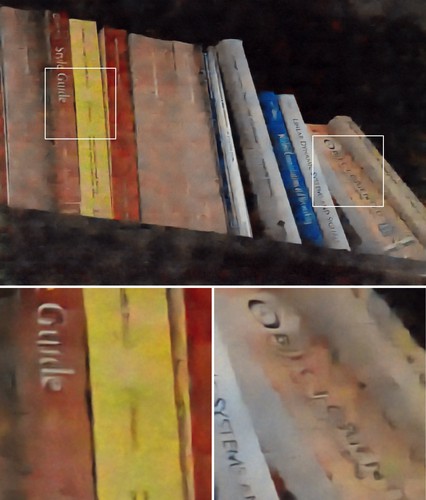}}
\subfloat{
\includegraphics[width=0.18\linewidth]{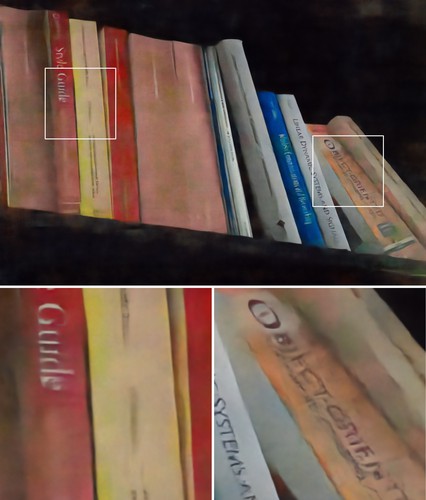}}
\subfloat{
\includegraphics[width=0.18\linewidth]{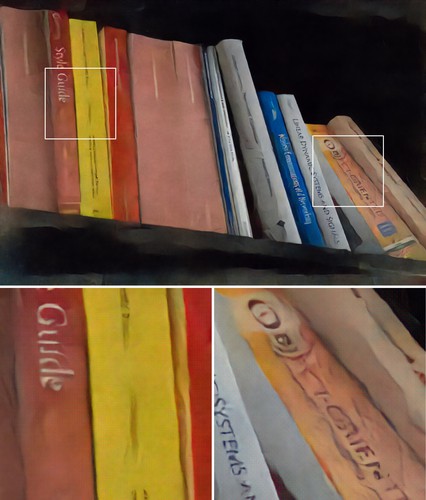}}
\subfloat{
\includegraphics[width=0.18\linewidth]{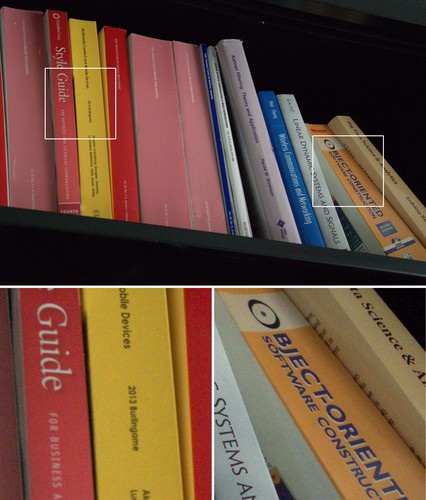}}

\subfloat{
\includegraphics[width=0.18\linewidth]{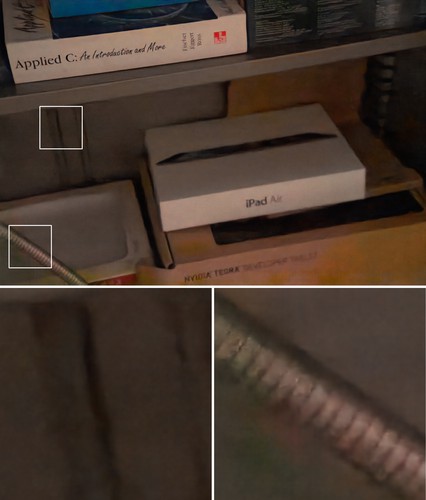}}
\subfloat{
\includegraphics[width=0.18\linewidth]{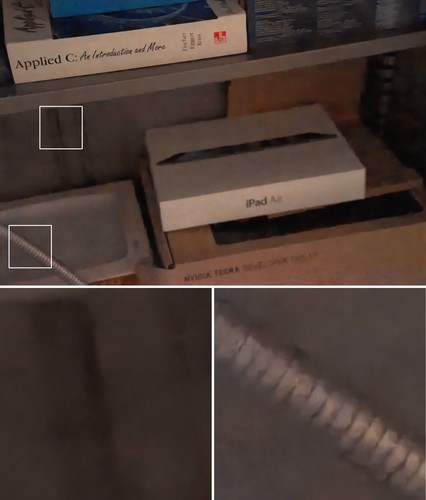}}
\subfloat{
\includegraphics[width=0.18\linewidth]{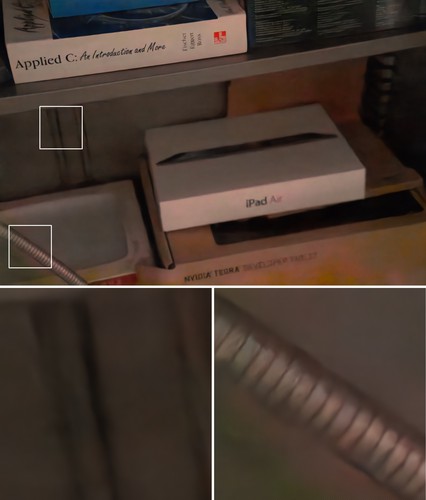}}
\subfloat{
\includegraphics[width=0.18\linewidth]{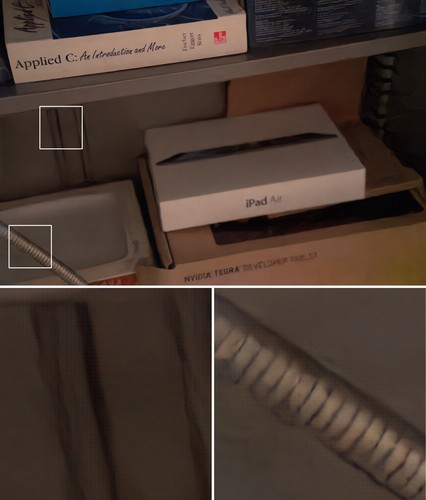}}
\subfloat{
\includegraphics[width=0.18\linewidth]{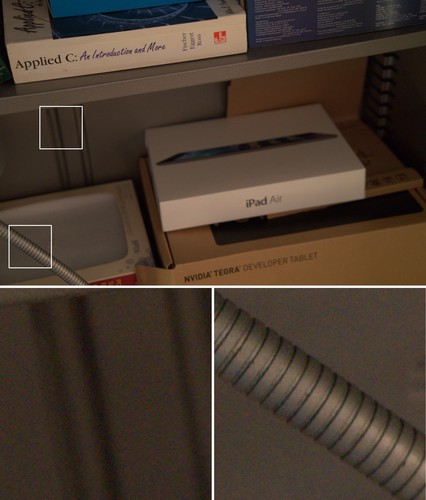}}

\setcounter{subfigure}{0}
\captionsetup[subfigure]{farskip=1pt,captionskip=1pt}

\subfloat[SID \cite{Chen2018}]{
\includegraphics[width=0.18\linewidth]{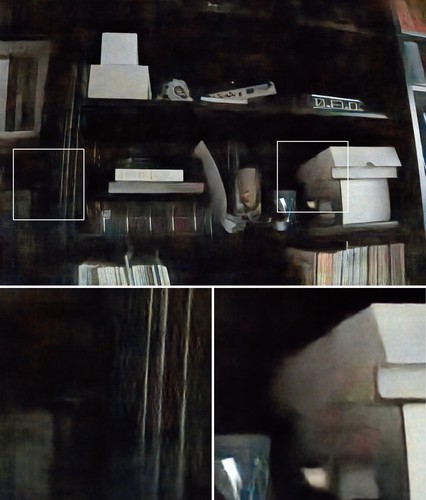}}
\subfloat[Maharjan et al. \cite{Maharjan2019a}]{
\includegraphics[width=0.18\linewidth]{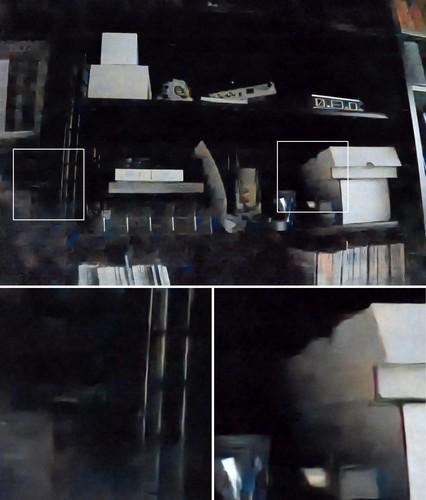}}
\subfloat[Zamir et al. \cite{Zamir2019a}]{
\includegraphics[width=0.18\linewidth]{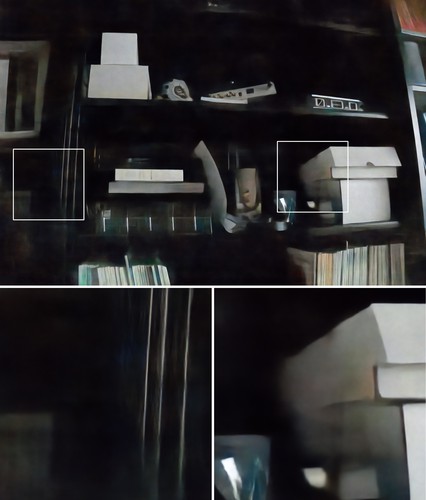}}
\subfloat[Ours \textit{(S)}]{
\includegraphics[width=0.18\linewidth]{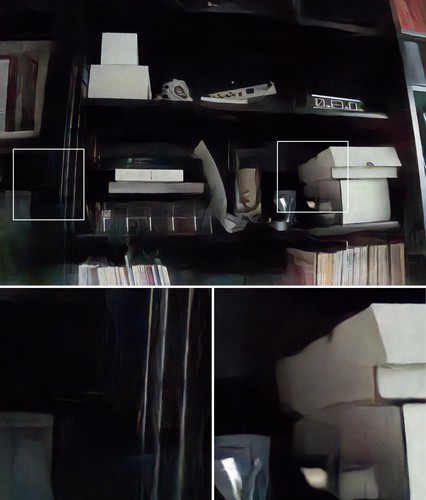}}
\subfloat[Ground truth]{
\includegraphics[width=0.18\linewidth]{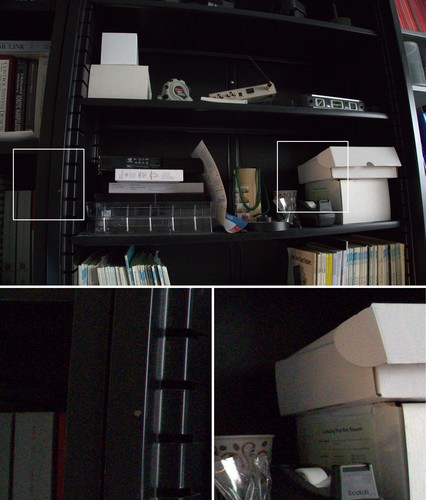}}
\caption{Qualitative comparison of our coarse-to-fine single image \textit{(S)} method for enhancing extremely low-light images, compared against the state-of-the-art models that also process single image. From top to the bottom row, the amplification ratios are $\times$250, $\times$100 and $\times$250, respectively.}
\label{fig:compare_single}
\end{figure*}

\begin{figure*}[!thb]
    \centering

\captionsetup[subfigure]{labelformat=empty,farskip=1pt,captionskip=1pt}
    
\subfloat{
\includegraphics[width=0.144\linewidth]{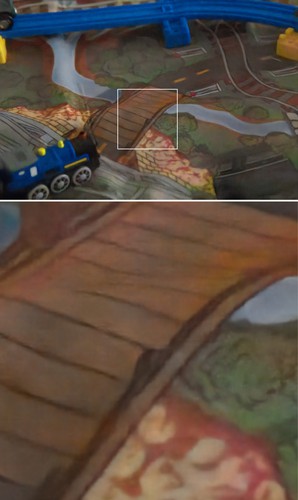}}
\subfloat{
\includegraphics[width=0.144\linewidth]{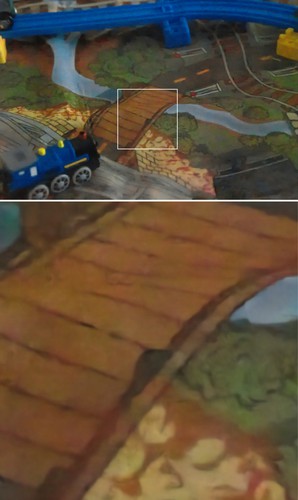}}
\subfloat{
\includegraphics[width=0.144\linewidth]{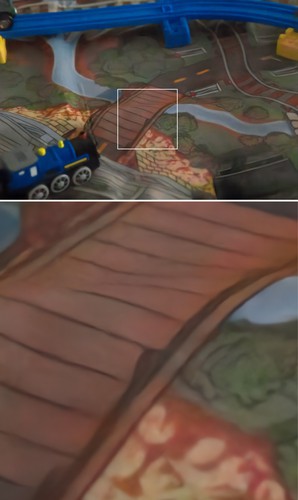}}
\subfloat{
\includegraphics[width=0.144\linewidth]{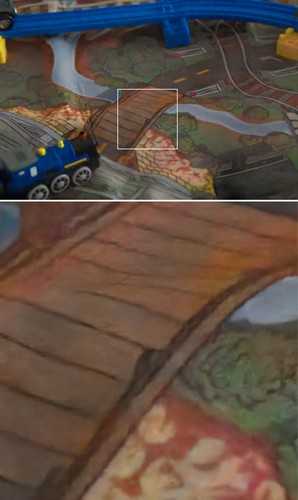}}
\subfloat{
\includegraphics[width=0.144\linewidth]{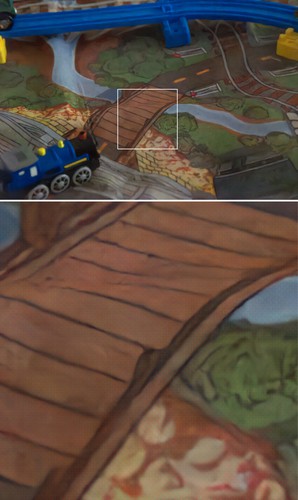}}
\subfloat{
\includegraphics[width=0.144\linewidth]{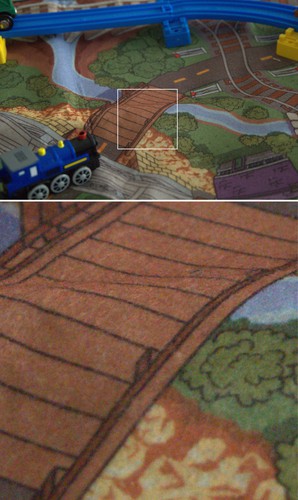}}

\subfloat{
\includegraphics[width=0.144\linewidth]{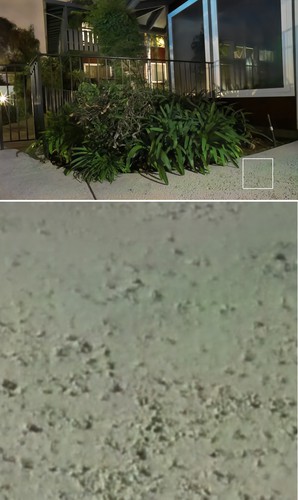}}
\subfloat{
\includegraphics[width=0.144\linewidth]{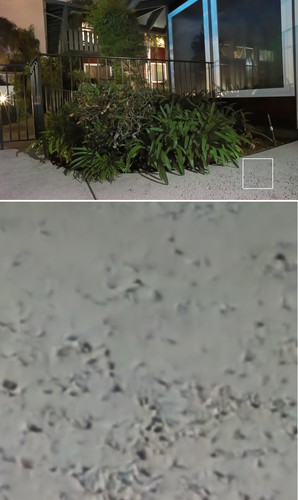}}
\subfloat{
\includegraphics[width=0.144\linewidth]{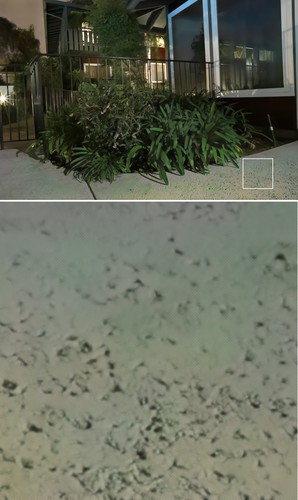}}
\subfloat{
\includegraphics[width=0.144\linewidth]{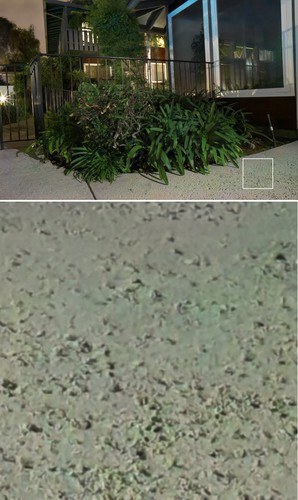}}
\subfloat{
\includegraphics[width=0.144\linewidth]{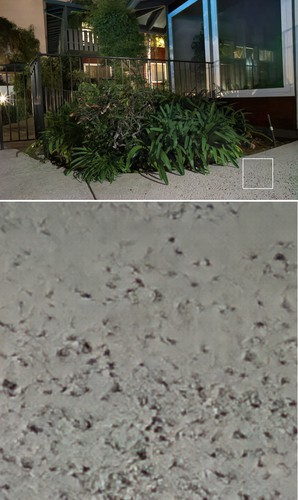}}
\subfloat{
\includegraphics[width=0.144\linewidth]{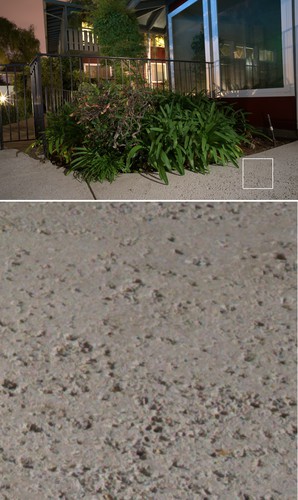}}

\captionsetup[subfigure]{farskip=1pt,captionskip=1pt}

\setcounter{subfigure}{0}
\captionsetup[subfigure]{farskip=1pt,captionskip=1pt}

\subfloat[SID \textit{(E)} \cite{Chen2018}]{
\includegraphics[width=0.144\linewidth]{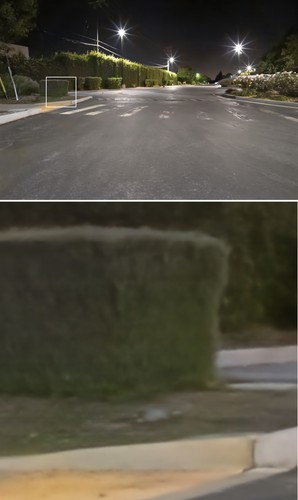}}
\subfloat[Maharjan et al. \textit{(E)} \cite{Maharjan2019a}]{
\includegraphics[width=0.144\linewidth]{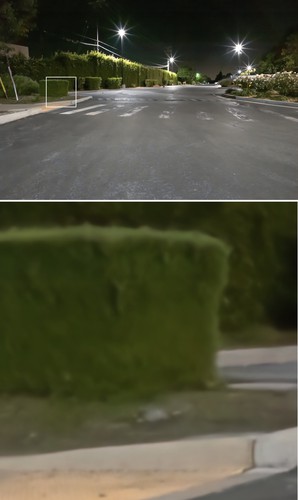}}
\subfloat[Zamir et al. \textit{(E)} \cite{Zamir2019a}]{
\includegraphics[width=0.144\linewidth]{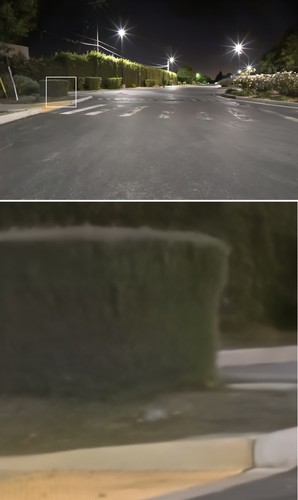}}
\subfloat[Ma et al. \textit{(B)} \cite{Ma2020}]{
\includegraphics[width=0.144\linewidth]{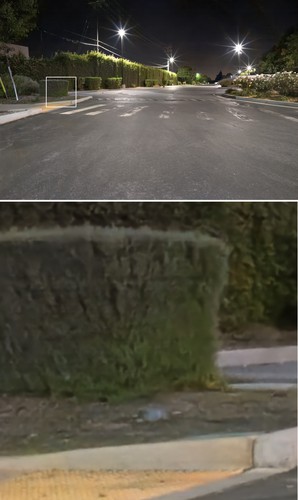}}
\subfloat[Ours \textit{(B)}]{
\includegraphics[width=0.144\linewidth]{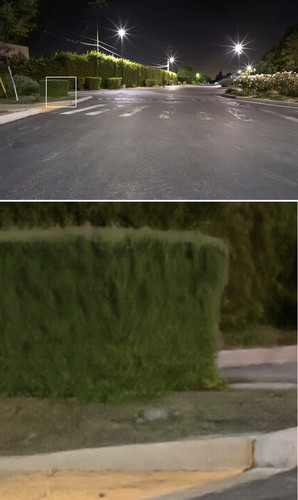}}
\subfloat[Ground truth]{
\includegraphics[width=0.144\linewidth]{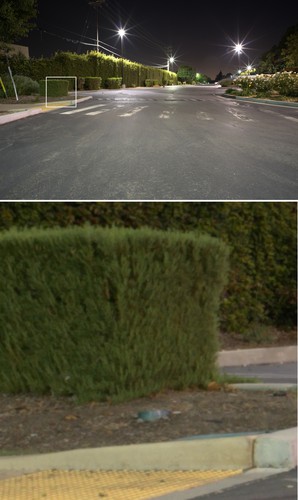}}

\caption{Qualitative comparison of our burst \textit{(B)} model for enhancing extremely low-light images, compared against the burst model by Ma et al.~\cite{Ma2020} and the ensemble versions \textit{(E)} of the single image state-of-the-art models. From top to the bottom row, the amplification ratios are $\times$100, $\times$300 and $\times$300 respectively.}
\label{fig:cmp_burst}
\end{figure*}

\textbf{Implementation Details.} To generate our training data, we extracted 512$\times$512 pixels random patches for each input image and also generated their downsampled versions with half resolution (obtained by bilinear interpolation). Hence, the input patch sizes for the coarse and fine networks are 256$\times$256 and 512$\times$512 pixels, respectively. We follow the same preprocessing steps for raw data as in~\cite{Chen2018} by packing raw array into channels, subtracting black level, and scaling the data with the given amplification ratio. We first trained the coarse network $F_c$ by using Adam optimizer with a learning rate of $10^{-4}$ for 2000 epochs and $10^{-5}$ for 2000 epochs. Then, the fine network $F_f$ was trained with the same hyperparameters without fixing the parameters of the coarse network. Finally, we trained the set-based network $F_s$ for 1000 epochs by initializing its weights from the fine network. During the training of $F_s$, we randomly chose the number of burst input frames between 1 and 8. We trained both of our models by using a hybrid loss that consists of the pixel-wise $L_\text{1}$ and the contextual $L_\text{CX}$ loss functions\footnote{In our experiments, we observed that the contextual loss $L_\text{CX}$ works consistently better than the perceptual loss $L_\text{P}$.}. For the contextual loss, we used \texttt{conv3\_2} and \texttt{conv4\_2} layers of the VGG-19 network. We implemented our model with Tensorflow library on an NVIDIA GeForce GTX 1080 Ti GPU. Training our model lasted about 4 days.

\section{Experimental Evaluation}
\subsection{Dataset}
 Obtaining long-exposure images is practically difficult but they can serve as ground truth images if the low-light scenes are static. We train and evaluate our models on the SID dataset~\cite{Chen2018}, which consists of short-exposure burst raw images taken under extremely dark indoor (0.2-5 lux) or outdoor (0.03-0.3 lux) scenes. These images are acquired with three different exposure times of 1/10, 1/25 and 1/30 sec, where the corresponding reference images are obtained with 10 seconds or 30 seconds exposures depending on the scene. We evaluate the performance of our models on the Sony and Fuji subsets. While the first one contains 161, 36 and 93 distinct burst sequences for training/validation/testing splits, the latter consists of 134, 38, and 93 sequences for training, validation and testing, respectively. The number of burst frames varies from 2 to 10 for each distinct scene. The burst images are totally aligned as they are captured with a tripod. The total number of images in this dataset is 5094, including the burst frames. Moreover, the images are categorized into three groups based on their amplification ratios ($\times$100, $\times$250, $\times$300), measured as the ratio between the exposure times of the dark input image and the long-exposure \mbox{ground truth}. In addition to the SID dataset, we also train and evaluate our model on the Dark Raw Video (DRV) dataset~\cite{chen2019seeing}. To our interest, this dataset contains both static videos with long exposure ground truth frames as well as dynamic videos with camera motion and/or moving objects. The videos are captured at 20 fps in dark (0.5-5 lux) and each is approximately 5.5 seconds long. There are a total of 128, 25 and 49 static videos for training, validation and testing split, respectively, and an additional set of 23 dynamic videos to qualitatively evaluate robustness to large motions.

\subsection{Competing Approaches}
We compare our models with four state-of-the-art methods, SID~\cite{Chen2018}, Maharjan et al.~\cite{Maharjan2019a}, Ma et al.~\cite{Ma2020} and Zamir et al.~\cite{Zamir2019a}. In our experiments, we used the pre-trained models provided by the authors of~\cite{Chen2018} and \cite{Maharjan2019a}, and our implementations of the methods in~\cite{Ma2020} and \cite{Zamir2019a} as their models are not publicly available. For Zamir et al.~\cite{Zamir2019a}, we trained the U-Net model with the hybrid loss including pixel-wise $L_1$ and \mbox{MS-SSIM} losses and the perceptual loss $L_\text{P}$ for 4000 epochs. For the burst-based model by Ma et al.~\cite{Ma2020}, we implemented a recurrent \mbox{U-Net} architecture, where the concatenated features from the previous frame, the single image model and the previous layer are fed to each convolution block of the network. We trained the model for 1000 epochs fixing the parameters of the single image network. Among these approaches, only the method by Ma et al.~\cite{Ma2020} processes a burst of images at once. Hence, for a fair comparison with the single image models, we process each burst image independently via each model, take the average of these enhanced outputs as the final result, and report the predictions of these ensemble models accordingly. We also compare our model with the Seeing Motion in the Dark (SMID) method of Chen et al. on the DRV dataset~\cite{chen2019seeing}.

\subsection{Evaluation Metrics}
We employ the popular peak signal-to-noise ratio (PSNR) and structural similarity index (SSIM) metrics, and also two perceptual image quality metrics, namely learned perceptual image patch similarity (LPIPS)~\cite{zhang2018unreasonable} and perceptual image-error assessment through pairwise preference (PieAPP)~\cite{prashnani2018pieapp}. These perceptual metrics can be used to quantify the natural distortion of images such as noise and blur as well as CNN-based distortions. We also employ perceptual index (PI)~\cite{blau20182018}, a recently proposed no-reference  image quality metric\footnote{We want to note that PI metric was originally suggested for evaluating the perceptual quality of super-resolution methods by taking into account traditional distortions like blur and noise.}.

\subsection{Experimental Results}
We first analyze the effectiveness of our coarse-to-fine strategy, and the performance gains achieved over the existing single image models. Fig.~\ref{fig:compare_single} shows visual comparison of our single image model against the state-of-the-art~\cite{Chen2018, Maharjan2019a, Zamir2019a}. For the first image, the color of the books and the details of texts contained on the spines are better recovered by our model. For the second image, the fine details are more visible and the edges are sharper (e.g. the lines on the wall and the cable) in our result. For the third image, our model greatly reduces the noise in the dark regions. Moreover, it is apparent that our approach preserves the edges better. Table~\ref{table:single_image_sony_fuji} shows quantitative analysis of our single image model on the SID Sony and SID Fuji datasets. Overall, our model outperforms the state-of-the-art in terms of PSNR, LPIPS, and PieAPP on the Sony subset. On the Fuji subset, we obtain the best results in terms of PSNR, SSIM, and PieAPP.

\begin{table*}[!t]
\caption{Performance comparison of single image models on the Sony and Fuji subsets of the SID dataset for different amplification ratios, with the best performing model highlighted with a bold typeface.}
\begin{center}
\resizebox{0.725\linewidth}{!} {
\begin{tabular}{c@{$\;\,$}lc@{$\;\,$}c@{$\;\,$}c@{$\;\,$}c@{$\;\,$}c@{$\qquad$}c@{$\;\,$}c@{$\;\,$}c@{$\;\,$}c@{$\;\,$}c}
\toprule
& & \multicolumn{5}{c}{Sony} & \multicolumn{5}{c}{Fuji}\\
\midrule
Ratio & Method & PSNR$\uparrow$ & SSIM$\uparrow$ & LPIPS$\downarrow$ & PieAPP$\downarrow$ & PI$\downarrow$ & PSNR$\uparrow$ & SSIM$\uparrow$ & LPIPS$\downarrow$ & PieAPP$\downarrow$ & PI$\downarrow$ \\
\midrule 
    {\multirow{4}{*}{\rotatebox[origin=c]{90}{$\times$100}}} & SID \cite{Chen2018} & 30.087 & 0.904 & 0.450 & 1.427 & \textbf{4.320} & 28.133 & 0.872 & 0.529 & 1.997 & \textbf{3.307} \\
    & Maharjan et al. \cite{Maharjan2019a} & 30.535 & \textbf{0.906} & 0.448  & 1.250 & 4.481 & $-$ & $-$ & $-$ & $-$ & $-$\\
    & Zamir et al. \cite{Zamir2019a} & 29.922 &  0.895 & 0.465 & 1.310 &  4.518 & 28.111 & 0.864 & \textbf{0.431} & 1.782 &  5.211\\
    & Ours & \textbf{31.178} & 0.905 & \textbf{0.285} & \textbf{1.038} & 4.546 & \textbf{29.651} & \textbf{0.877} & 0.465 & \textbf{1.607} & 3.561\\
\midrule 
    {\multirow{4}{*}{\rotatebox[origin=c]{90}{$\times$250}}} & SID \cite{Chen2018} & 28.428 & 0.887 & 0.482 & 1.601 & \textbf{4.577} & 26.477 & 0.824 & 0.611 & 1.960 & \textbf{3.885}\\
    & Maharjan et al. \cite{Maharjan2019a} & 28.787 &  0.888 & 0.488  & 1.443 & 4.961  & $-$ & $-$ & $-$ & $-$ & $-$ \\
    & Zamir et al. \cite{Zamir2019a} & 28.254 & 0.878 & 0.462 & 1.462 & 4.956 & 26.581 & 0.817 & \textbf{0.501} & 1.767 & 7.613\\
    & Ours & \textbf{29.337} & \textbf{0.888} & \textbf{0.315} & \textbf{1.053} & 4.906 & \textbf{27.535} & \textbf{0.830} & 0.546 & \textbf{1.601} & 4.189\\
\midrule
    {\multirow{4}{*}{\rotatebox[origin=c]{90}{$\times$300}}} & SID \cite{Chen2018} & 28.528 & 0.870 & 0.507 & 1.644 & \textbf{4.107} & 25.509 & 0.799 & 0.647 & 2.019 & \textbf{3.801} \\
    & Maharjan et al. \cite{Maharjan2019a} & 28.382 &  0.868 & 0.516  & 1.645 & 4.523  & $-$ & $-$ & $-$ & $-$ & $-$ \\
    & Zamir et al. \cite{Zamir2019a} & 28.441 &  0.860 & 0.494 & 1.520 & 4.479 & 25.394 &  0.791 & \textbf{0.555} & 1.768 & 7.524\\
    & Ours & \textbf{29.018} & \textbf{0.870} & \textbf{0.347} & \textbf{1.155} & 4.349 & \textbf{26.583} & \textbf{0.817} & 0.587 & \textbf{1.609} & 3.928\\
\midrule
    {\multirow{4}{*}{\rotatebox[origin=c]{90}{All}}} & SID \cite{Chen2018} & 28.976 & 0.886 & 0.482 & 1.564 & \textbf{4.319} & 26.940 & 0.838 & 0.585 & 1.992 & \textbf{3.606} \\
    & Maharjan et al. \cite{Maharjan2019a} & 29.167 &  0.886 & 0.487  & 1.462 & 4.646  & $-$ & $-$ & $-$ & $-$ & $-$\\
    & Zamir et al. \cite{Zamir2019a} & 28.838 &  0.876 & 0.465 & 1.437 & 4.639 & 26.930 & 0.831 & \textbf{0.485} & 1.767 & 6.524 \\
    & Ours \textit{(S)} & \textbf{29.780} & \textbf{0.886} & \textbf{0.318} & \textbf{1.088} & 4.581 & \textbf{28.204} & \textbf{0.847} & 0.521 & \textbf{1.606} & 3.841\\
\bottomrule
\end{tabular}
}
\end{center}
\label{table:single_image_sony_fuji}
\end{table*}

\begin{table*}[!t]
\caption{Performance comparison of burst \textit{(B)} and ensemble \textit{(E)} models on the Sony and Fuji subsets of the SID dataset for different amplification ratios, with the best performing model highlighted with a bold typeface.}
\begin{center}
\resizebox{0.725\linewidth}{!} {
\begin{tabular}{c@{$\;\,$}l@{$\;\,$}c@{$\;\,$}c@{$\;\,$}c@{$\;\,$}c@{$\;\,$}c@{$\qquad$}c@{$\;\,$}c@{$\;\,$}c@{$\;\,$}c@{$\;\,$}c}
\toprule
& & \multicolumn{5}{c}{Sony} & \multicolumn{5}{c}{Fuji}\\
\midrule
Ratio & Method & PSNR$\uparrow$ & SSIM$\uparrow$ & LPIPS$\downarrow$ & PieAPP$\downarrow$ & PI$\downarrow$ & PSNR$\uparrow$ & SSIM$\uparrow$ & LPIPS$\downarrow$ & PieAPP$\downarrow$ & PI$\downarrow$ \\
\midrule
    \multirow{5}{*}{\rotatebox[origin=c]{90}{$\times$100}} & SID \textit{(E)} \cite{Chen2018} & 30.361 & 0.908 & 0.447 & 1.441 & 4.686 & 28.520 & 0.876 & 0.523 & 1.957 & 3.574\\
    & Maharjan et al. \textit{(E)} \cite{Maharjan2019a} & 30.833 &  \textbf{0.909} & 0.445  & 1.324 & 4.863  & - & - & - & - & - \\
    & Zamir et al. \textit{(E)} \cite{Zamir2019a} & 30.120 &  0.898 & 0.430 &  1.335 & 4.776 & 28.458 & 0.868 & \textbf{0.431} & 1.782 & 5.408\vspace{0.03in}\\
    & Ma et al. \textit{(B)} \cite{Ma2020} & 30.429 &  0.908 & 0.423 & 1.312 & \textbf{4.295} & 28.890 & 0.879 & 0.503 & 1.808 & \textbf{3.416}\\
    & Ours \textit{(B)} & \textbf{31.330} & 0.906 & \textbf{0.279} & \textbf{1.027} & 4.493 & \textbf{30.054} & \textbf{0.881} & 0.445 & \textbf{1.580} & 3.453\\
\midrule
    \multirow{5}{*}{\rotatebox[origin=c]{90}{$\times$250}} & SID \textit{(E)} \cite{Chen2018} & 28.915 & 0.893 & 0.480 & 1.622 & 5.313 & 27.117 & 0.834 & 0.596 & 1.941 & 4.610\\
    & Maharjan et al. \textit{(E)} \cite{Maharjan2019a} & 29.289 &  0.893 & 0.480  & 1.525 & 5.609  & - & - & - & - & - \\
    & Zamir et al. \textit{(E)} \cite{Zamir2019a} & 28.630 &  0.882 & 0.454 &  1.495 & 5.406 & 27.094 & 0.823 & \textbf{0.493} & 1.767 & 8.095\vspace{0.03in}\\
    & Ma et al. \textit{(B)}\cite{Ma2020} & 29.053 &  \textbf{0.896} & 0.470 & 1.517 & \textbf{4.429} & 27.135 & 0.834 & 0.611 & 1.781 & \textbf{3.881} \\
    & Ours \textit{(B)} &\textbf{29.661} & 0.892 & \textbf{0.303} & \textbf{1.041} &  4.791 & \textbf{27.925} & \textbf{0.835} & 0.530 & \textbf{1.573} & 3.958\\
\midrule
     \multirow{5}{*}{\rotatebox[origin=c]{90}{$\times$300}} & SID \textit{(E)} \cite{Chen2018} & 28.979 & 0.878 & 0.516 & 1.699 & 4.606 & 26.240 & 0.815 & 0.643 & 1.971 & 4.340\\
    & Maharjan et al. \textit{(E)} \cite{Maharjan2019a} & 28.783 &  0.875 & 0.520  & 1.744 & 5.003  & - & - & - & - & - \\
     & Zamir et al. \textit{(E)} \cite{Zamir2019a} & 28.750 &  0.866 & 0.500 &  1.581 & 4.805 & 25.977 & 0.802 & \textbf{0.553} & 1.768 & 8.159\vspace{0.03in}\\
    & Ma et al. \textit{(B)}\cite{Ma2020} & 29.078 &  \textbf{0.884} & 0.467 & 1.464 & \textbf{4.018} & 26.043 & 0.815 & 0.625 & 1.834 & 3.974 \\
    & Ours \textit{(B)} & \textbf{29.324} & 0.874 & \textbf{0.334} & \textbf{1.141} & 4.161 & \textbf{27.174} & \textbf{0.826} & 0.555 & \textbf{1.585} & \textbf{3.722}\\
\midrule
    \multirow{5}{*}{\rotatebox[origin=c]{90}{All}} & SID \textit{(E)} \cite{Chen2018} & 29.383 & 0.892 & 0.484 & 1.596 & 4.850 & 27.493 & 0.847 & 0.577 & 1.957 & 4.077\\
    & Maharjan et al. \textit{(E)} \cite{Maharjan2019a} & 29.568 &  0.891 & 0.485  & 1.548 & 5.148  & - & - & - & - & - \\
    & Zamir et al. \textit{(E)} \cite{Zamir2019a} & 29.132 &  0.881 & 0.462 &  1.480 & 4.983 & 27.387 & 0.837 & \textbf{0.482} & 1.773 & 6.922\vspace{0.03in}\\
    & Ma et al. \textit{(B)}\cite{Ma2020} & 29.485 &  \textbf{0.895} & 0.455 & 1.433 & \textbf{4.232} & 27.607 & 0.848 & 0.567 & 1.807 & 3.702 \\
    & Ours \textit{(B)} & \textbf{30.043} & 0.890 & \textbf{0.308} & \textbf{1.076} & 4.461 & \textbf{28.655} & \textbf{0.853} & 0.499 & \textbf{1.579} & \textbf{3.671}\\
\bottomrule
\end{tabular}
}
\end{center}
\label{table:burst-sony-burst}
\end{table*}

Fig.~\ref{fig:cmp_burst} presents some visual results of our burst model, along with a performance comparison to the burst method of~\cite{Ma2020} and the ensemble versions of the single image methods~\cite{Chen2018,Maharjan2019a,Zamir2019a}. As evident from the zoomed-in regions, our permutation-invariant CNN model can produce enhancement results with much sharper and well restored texture details. On the other hand, the ensemble methods all suffer from over-smoothing of the fine-scale details such as the thin lines on the mat and the textured regions like the ground or green bush. The burst method of~\cite{Ma2020} does relatively better but its outputs are of low contrast. Table~\ref{table:burst-sony-burst} clearly demonstrates the effectiveness of our approach that it achieves the best overall scores on the Sony and Fuji subsets of the SID dataset.

\begin{figure}[!t]
\centering
\begin{tabular}{c@{$\;$}c}
     \includegraphics[width=0.48\linewidth]{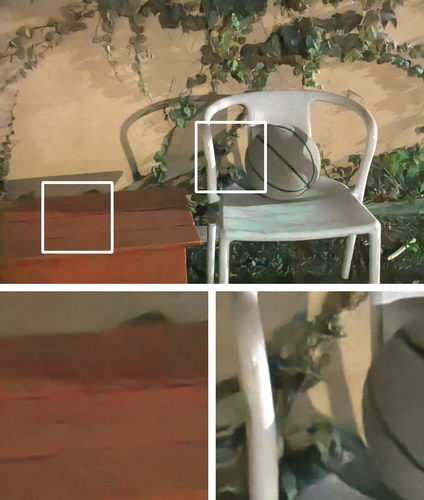} &  
     \includegraphics[width=0.48\linewidth]{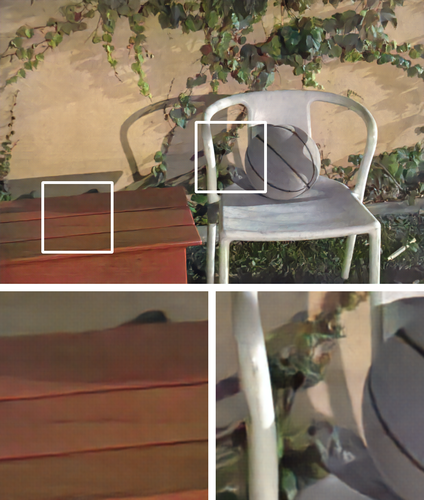} \\
     {\footnotesize SMID~\cite{chen2019seeing}} &
     {\footnotesize Ours}
\end{tabular}
\caption{Qualitative comparison of our burst method for enhancing extremely low-light static videos, compared against the SMID method of Chen et al.~\cite{chen2019seeing}.}
\label{fig:motion_static}
\end{figure}

\begin{figure}[!t]
\centering
\begin{tabular}{c@{\;}c@{\;}c}
\includegraphics[width=0.32\linewidth]{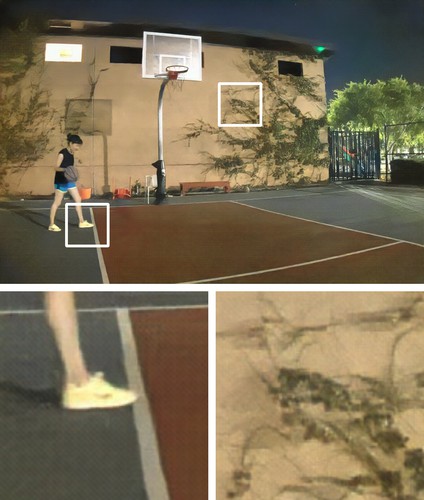} & 
\includegraphics[width=0.32\linewidth]{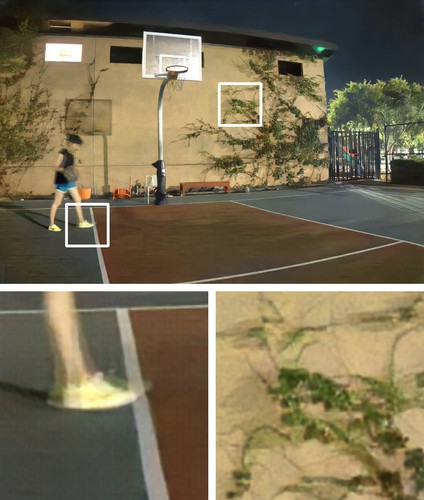} &
\begin{overpic}[width=0.32\linewidth]{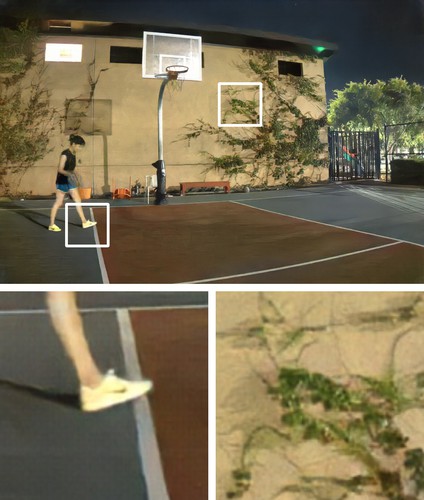}
 \put(0,79){\includegraphics[width=0.06\textwidth]{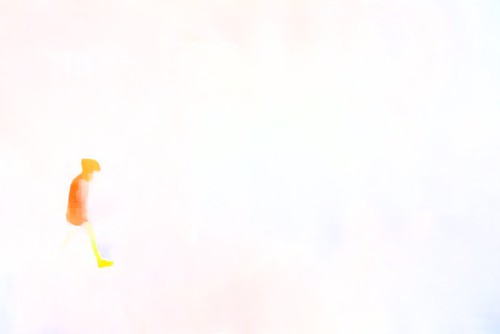}}
\end{overpic}\\
{\footnotesize SMID~\cite{chen2019seeing}} &
{\footnotesize Ours w/o motion comp.} & 
{\footnotesize Ours}
\end{tabular}    
\caption{Qualitative comparison of our burst method for enhancing extremely low-light dynamic videos, compared against the SMID method of Chen et al.~\cite{chen2019seeing}.}
\label{fig:motion_dynamic}
\end{figure}

In Fig.~\ref{fig:motion_static}, we show a comparison of our method to SMID~\cite{chen2019seeing} on a video from the static test set of DRV and the difference can clearly be seen. Our approach gives better result in terms of clarity and sharpness.  Table~\ref{table:motion} presents a quantitative evaluation on this dataset showing that our model outperforms Chen et al.'s SMID~\cite{chen2019seeing} method on both PSNR, SSIM, and LPIPS metrics. In Fig.~\ref{fig:motion_dynamic}, we provide a qualitative result for a dynamic sequence from DRV and compare our approach with and without motion compensation against SMID. Both our burst model and SMID can handle large motion, but our approach can capture more realistic texture details with more vivid colors. To further validate that our approach can be also effectively applied to dynamic scenes, we conducted a user study on 22 dynamic test sequences from DRV since their long-exposure ground truths are not available. We showed video patches of size $600 \times 450$ pixels side-by-side\footnote{Here, in addition to this sampling procedure, we also slowed down the video patches to 7 fps to better highlight the differences in the outputs of the methods.}, which are respectively sampled from our results and those of SMID, and asked 15 participants to identify the ones that they think have better perceived quality. We found that $58\%$ of the subjects preferred our results against SMID. Moreover, in terms of no-reference quality metric PI, our model has a better score (3.762) compared to SMID (3.809).

\begin{table}[!t]
\caption{Performance comparison with the Seeing Motion in the Dark on DRV dataset.}
\begin{center}
\begin{tabular}{l@{$\quad$}c@{$\quad$}c@{$\quad$}c@{$\quad$}c@{$\quad$}c}
\toprule
Method & PSNR$\uparrow$ & SSIM$\uparrow$ & LPIPS$\downarrow$ \\
\midrule
SMID \cite{chen2019seeing} & 28.474 & 0.906 & 0.357 \\
Ours \textit{(S)} & 28.671 & 0.910 & 0.345 \\
Ours \textit{(B)} (3 frames) & 28.924 & 0.915 & 0.336 \\
Ours \textit{(B)} (8 frames) & \textbf{29.104} & \textbf{0.918} & \textbf{0.329}\\
\bottomrule
\end{tabular}
\end{center}
\label{table:motion}
\end{table}

In Table~\ref{table:runtime}, we report the runtime performances of our single image and burst models in comparison with other competing methods. In particular, we measure the time taken to process a single image and also a burst of 4 images. Our experiments are conducted on a machine with an NVIDIA GeForce GTX 1080 Ti 11GB graphics card using 4256$\times$2848 pixels images. For single image enhancement, our single image model is a bit slower than SID~\cite{Chen2018} and Zamir et al~\cite{Zamir2019a} due to its multi-scale architecture, though it gives better enhancement results as discussed before. For burst enhancement, our model achieves the second-best runtime performance, with 1.229 sec for a burst size of 4. This clearly demonstrates the advantage of having a shared decoder to process burst features, contrary to the competing approaches. We additionally report the runtime of our burst model to enhance a burst of 8 frames. As can be seen, the increase in the runtime is not linear in the number of processed images. We only observe 1.7$\times$ increase when the burst size is doubled from 4 to 8. We note that for the case of the burst size of 8, we were unable to report runtimes of the competing models as enhancing these frames within a single batch by these models exceed the limits of our GPU memory.

\begin{table}[!t]
\caption{Runtime analysis for single image and ensemble/burst models. Running times are in seconds.}
\begin{center}
    \begin{tabular}{l@{$\qquad$}c@{$\qquad$}c@{$\qquad$}c}
    \toprule
        Method & 1 frame & 4 frames & 8 frames\\
    \midrule
        SID \cite{Chen2018} & 0.307 & 1.069 & -- \\
        Maharjan et al. \cite{Maharjan2019a} &  2.287 & 3.045  & -- \\
        Zamir et al.  \cite{Zamir2019a} & 0.307 & 1.069 & -- \\
        Ma et al. \cite{Ma2020} & -- & 1.604 & -- \\
        Ours \textit{(S)} & 0.566 & 2.152 & -- \\
        Ours \textit{(B)} & 0.566 &  1.229 & 2.113\\
    \bottomrule
    \end{tabular}
\end{center}
\label{table:runtime}
\end{table}

\begin{figure}[!t]
\centering
\begin{tabular}{c@{$\;$}c}
     \includegraphics[width=0.48\linewidth]{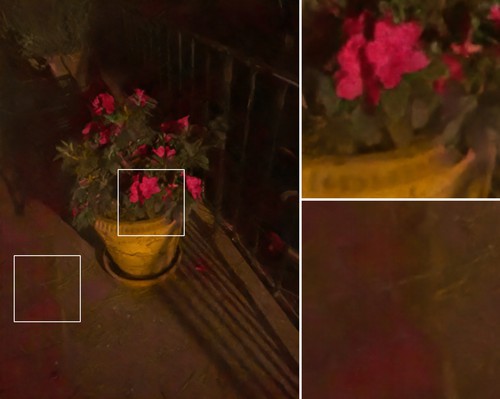} &  
     \includegraphics[width=0.48\linewidth]{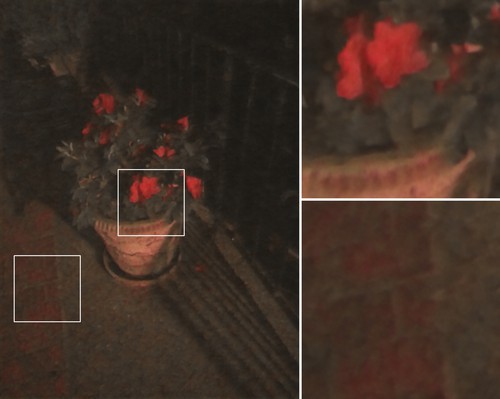} \vspace{-0.1cm}\\\vspace{0.1cm}
     {\footnotesize SID~\cite{Chen2018}} &
     {\footnotesize Maharjan et al.~\cite{Maharjan2019a}} \\
     \includegraphics[width=0.48\linewidth]{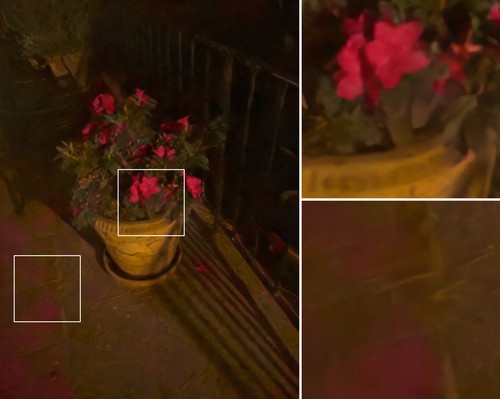} &
     \includegraphics[width=0.48\linewidth]{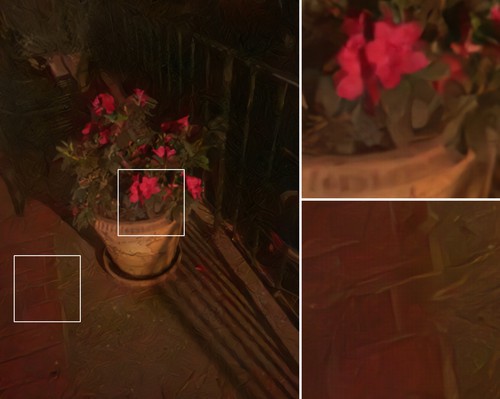} \vspace{-0.1cm}\\
     {\footnotesize Zamir et al.~\cite{Zamir2019a}} 
     & 
     {\footnotesize Ours (single)}
\end{tabular}
\caption{Enhancement results of a raw image captured by an iPhone 6s using 1/20 sec exposure time and 400~ISO. Our proposed single image enhancement model provides better noise reduction with more structural details, in comparison to the prior approaches.}
\label{fig:iphone1}
\end{figure}

\begin{figure}[!t]
\centering
\subfloat[Traditional Pipeline\\\centering\textit{(Ensemble)}]{
\includegraphics[width=0.32\linewidth]{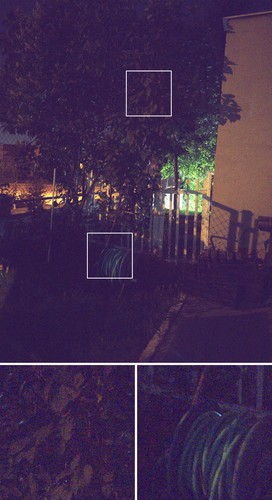}}
\subfloat[SID \textit{(Ensemble)}~\cite{Chen2018}]{
\includegraphics[width=0.32\linewidth]{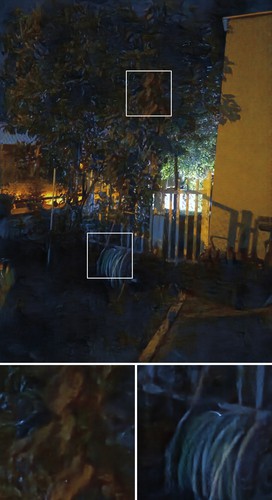}}
\subfloat[Ours \textit{(Burst)}]{
\includegraphics[width=0.32\linewidth]{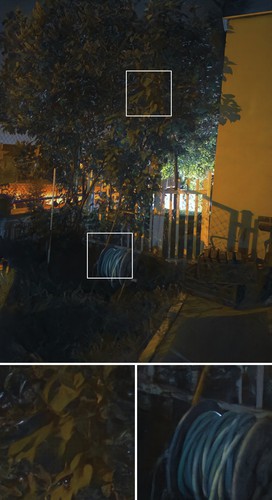}}

\caption{Enhancement results on a burst of 8 raw images taken with an iPhone SE with 1/10 sec exposure time and 400 ISO. Resulting images obtained by (a) averaging over the traditional pipeline, (b) averaging over the SID~\cite{Chen2018} predictions, (c) our burst model.}
\label{fig:iphone2}
\end{figure}

To show that our models can (partly) generalize to other camera sensors, in Fig.~\ref{fig:iphone1} and Fig.~\ref{fig:iphone2}, we present example outputs of our single and burst image models on extremely dark photos taken with cameras of an iPhone 6s and an iPhone SE, respectively\footnote{We used our models trained on the Sony subset of SID since iPhone cameras have a Bayer filter array similar to that of the Sony $\alpha$7S II camera used in collecting this subset.}. Once again, Fig.~\ref{fig:iphone1} demonstrates that our model reduces the noise better than the state-of-the-art models~\cite{Chen2018, Maharjan2019a, Zamir2019a}, while accurately improving the texture details of the flower and the leaves. Similarly, Fig.~\ref{fig:iphone2} shows the cross-sensor generalization capability of our burst model. Our method produces a better result than both the traditional camera pipeline\footnote{\url{https://github.com/letmaik/rawpy}} and SID~\cite{Chen2018} in that it recovers the details of the water hose and the leaves of the tree more accurately.

\subsection{Ablation Study}
To better understand the effect of each component of our proposed architecture and the loss functions and to evaluate the contribution of the burst size to the overall quality, we conducted an extensive series of ablation tests.

\textbf{Architectures}
In Table~\ref{table:ablation_architecture}, we show the effect of each component for our architecture trained with single frame and $L_1$ loss. \emph{U-Net} indicates a standard U-Net used in \cite{Chen2018}, \emph{Coarse-to-fine} refers to our multiscale approach, \emph{Residual} denotes the extension where residual blocks are used between the encoder and decoder, and finally, \emph{SE} corresponds to the case where squeeze-excitation layers are inserted into the residual blocks. As can be seen, our coarse-to-fine architecture improves all the scores. Moreover, these results are further improved with the use of residual blocks and the squeeze excitation layers.

\textbf{Losses.}
As mentioned before, the loss function we used to train our networks consists of two complementary loss terms. The first term is the pixel-wise $L_\text{1}$ loss which is used to improve the accuracy of reconstructing a long-exposure image. The second term, on the other hand, is comprised of the contextual $L_{\text{CX}}$ loss function, which is utilized to improve the perceived quality of the end result. 

\begin{figure}[!t]
    \centering
\subfloat{
\includegraphics[width=0.28\linewidth]{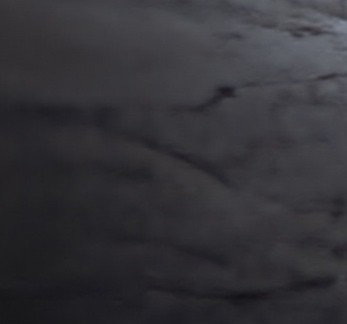}}
\subfloat{
\includegraphics[width=0.28\linewidth]{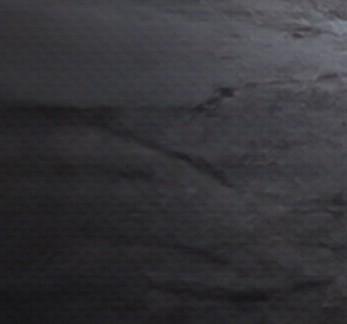}}
\subfloat{
\includegraphics[width=0.28\linewidth]{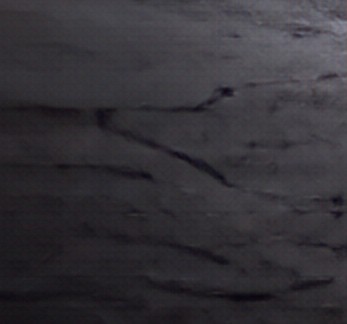}}
\vspace{-0.25cm}
\setcounter{subfigure}{0}
\subfloat[${L}_\text{1}$]{
\includegraphics[width=0.28\linewidth]{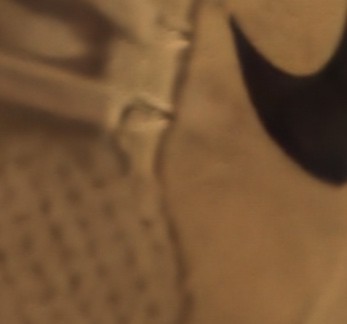}}
\subfloat[$L_\text{1} + L_\text{P}$]{
\includegraphics[width=0.28\linewidth]{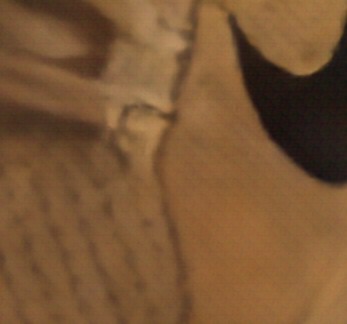}}
\subfloat[$L_\text{1} + L_\text{CX}$]{
\includegraphics[width=0.28\linewidth]{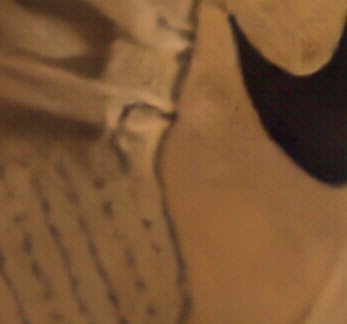}}

\caption{Results of our method with different loss functions. Combination of contextual loss and pixel-wise loss gives visually more pleasing results, as compared to using the pixel-wise loss together with and without the perceptual loss.}
\label{fig:loss}
\end{figure}

In Table~\ref{table:ablation_loss}, we quantify the effect of using the contextual loss, as opposed to the perceptual loss, in conjunction with the pixel-wise $L_\text{1}$ loss. First of all, the burst model trained with only $L_\text{1}$ loss results in higher PSNR and SSIM but relatively lower perceptual quality, which is in line with the previous observations~\cite{blau20182018, zhang2018unreasonable}. In that sense, adding either $L_{\text{P}}$ or $L_{\text{CX}}$ to our objective function provides a good tradeoff between pixel-wise and perceptual metrics. To inspect which one is better, we also qualitatively analyze the contribution of incorporating the perceptual loss $L_{\text{P}}$ or the contextual loss  $L_{\text{CX}}$. As demonstrated in Fig.~\ref{fig:loss}, either $L_{\text{P}}$ or $L_{\text{CX}}$ allows improving the perceived quality of the end-result. The resulting images have more realistic fine-scale details and texture while avoiding over-smoothing. To our interest, however, the network trained with the contextual loss tends to better recover the thin structures, especially at the darker regions, as compared to the others. 

\begin{table}[!t]
\caption{Comparison of the single-frame network architectures trained with $L_1$ loss.}
\begin{center}
\begin{tabular}{l@{$\quad$}c@{$\quad$}c@{$\quad$}c@{$\quad$}c@{$\quad$}c}
\toprule
Method & PSNR$\uparrow$ & SSIM$\uparrow$ & LPIPS$\downarrow$ \\
\midrule
U-Net & 28.976 & 0.886 & 0.482 \\
U-Net + Coarse-to-fine & 29.426 & 0.889 & 0.468 \\
U-Net + Coarse-to-fine + Residual & 29.812 & 0.890  & 0.466 \\
U-Net + Coarse-to-fine + Residual + SE & \textbf{29.939} & \textbf{0.892} & \textbf{0.466} \\
\bottomrule
\end{tabular}
\end{center}
\label{table:ablation_architecture}
\end{table}

\begin{table}
\caption{Effect of the loss functions on the performance of the proposed burst enhancement model.}
\begin{center}
\begin{tabular}{l@{$\;\,$}c@{$\;\,$}c@{$\;\,$}c@{$\;\,$}c@{$\;\,$}c}
\toprule
Method & PSNR$\uparrow$ & SSIM$\uparrow$ & LPIPS$\downarrow$ & PieAPP$\downarrow$ & PI$\downarrow$ \\
\midrule
    $L_\text{1}$ & \textbf{30.110} & \textbf{0.895} & 0.444 & 1.454 & \textbf{4.437} \\
    $L_\text{1}+L_\text{P}$ & 29.927 & 0.887 & \textbf{0.276} & 1.228 & 4.896 \\
    $L_\text{1}+L_\text{CX}$ & 30.043 & 0.890 & 0.308 & \textbf{1.076} & 4.461\\
\bottomrule
\end{tabular}
\end{center}
\label{table:ablation_loss}
\end{table}

\begin{figure}[!t]
\centering
\setcounter{subfigure}{0}
\captionsetup[subfigure]{labelformat=empty,farskip=1pt,captionskip=1pt}
\subfloat{
\includegraphics[width=0.84\linewidth]{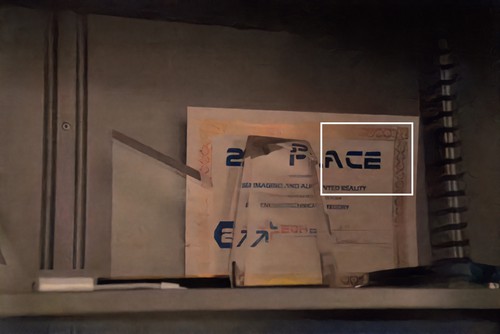}}\\
\subfloat[Single image]{
\includegraphics[width=0.20\linewidth]{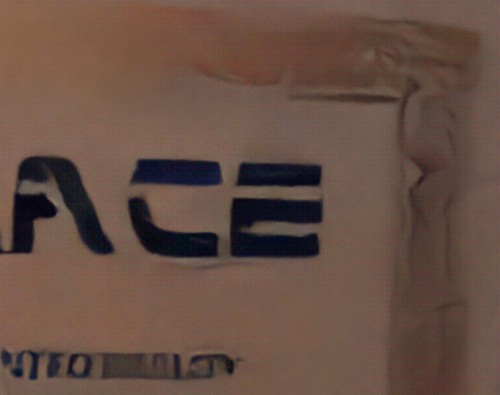}}
\subfloat[4 frames]{
\includegraphics[width=0.20\linewidth]{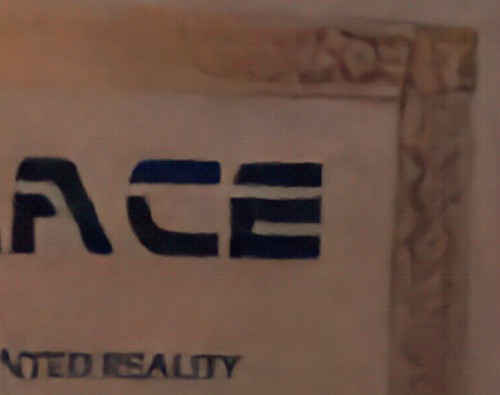}}
\subfloat[8 frames]{
\includegraphics[width=0.20\linewidth]{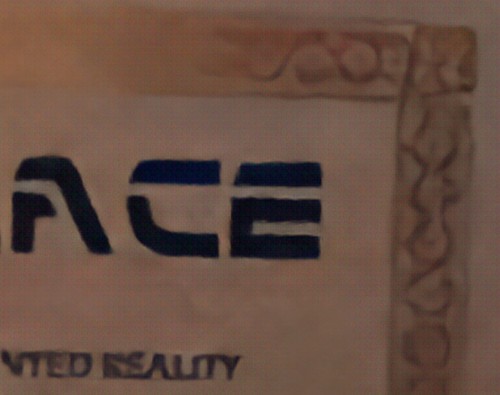}}
\subfloat[Ground truth]{
\includegraphics[width=0.20\linewidth]{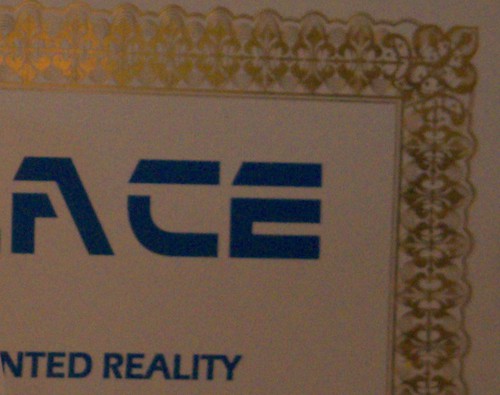}}
\caption{Effect of the burst size. As can be seen, as we increase the number of images in the burst sequence, the enhancement quality of our burst model improves further.}
\label{fig:num_frames}
\end{figure}

\begin{figure*}[!t]
\centering
\setcounter{subfigure}{0}
\captionsetup[subfigure]{labelformat=empty,farskip=1pt,captionskip=1pt}
\subfloat[Ours (ensemble) 0.754/27.001/0.470]{
\includegraphics[width=0.24\linewidth]{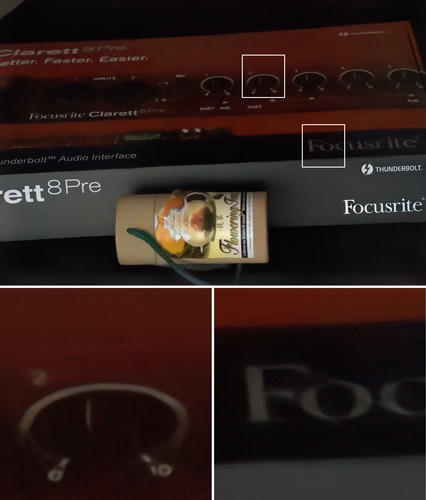}}
\subfloat[Ours (burst) 0.740/26.728/0.451]{
\includegraphics[width=0.24\linewidth]{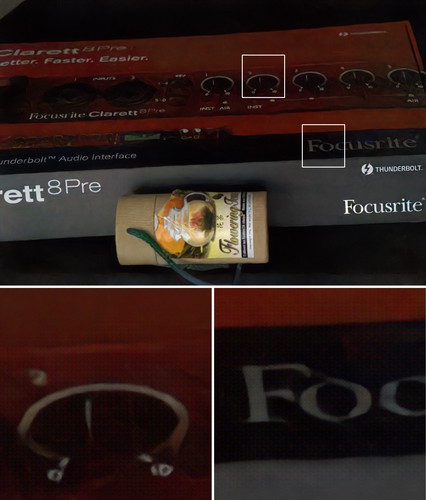}}
\subfloat[Ours (ensemble) 0.950/32.714/0.282]{
\includegraphics[width=0.24\linewidth]{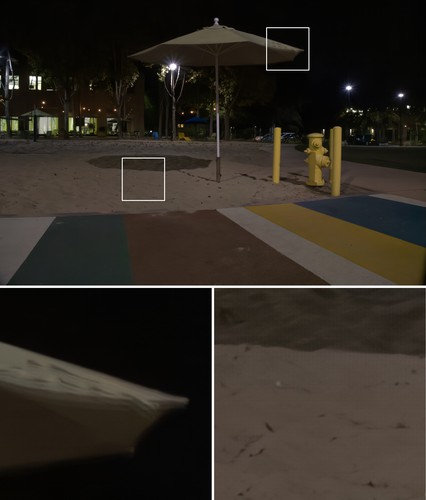}}
\subfloat[Ours (burst) 0.950/33.890/0.229]{
\includegraphics[width=0.24\linewidth]{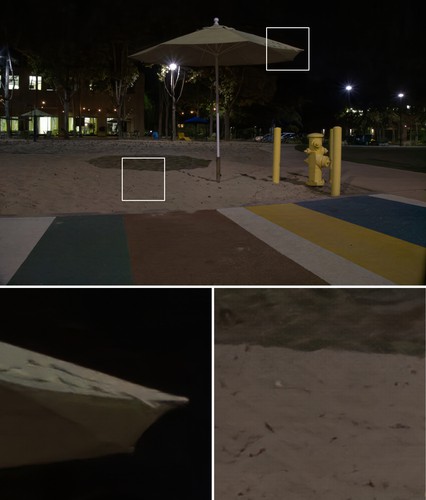}}
\caption{A comparison between our burst model and the ensemble version of our single image model for a burst size of 8 images. Our set-based approach, which performs fusion at the feature-level, gives perceptually better enhancement results. SSIM, PSNR, and LPIPS scores are also given for each result in this given order.}
\label{fig:ensemble}
\end{figure*}

\textbf{Burst Processing.}
In Fig.~\ref{fig:num_frames}, we analyze how the number of frames in the burst sequence affects the performance of our model. Here, we provide the results obtained with a single input image and the burst sizes of four and eight frames. As can be seen from the zoomed-in results, the output quality improves with an increasing number of the burst images, the method gets much better at preserving texture details and thin structures. Table~\ref{table:ablation_numframes} quantitatively shows that the results of our method improve with more images. In Fig.~\ref{fig:ensemble}, we also compare our (set-based) burst method with the ensemble of our single image model (i.e., processing each image in the burst separately and then taking the average of individual outputs). Fusing burst images at the feature level is evidently much more effective. Additionally, in Table~\ref{table:ablation_ensemble}, we quantitatively evaluate the performance of these alternative strategies on the SID and DRV datasets\footnote{The burst sizes for the images in the SID dataset vary between 2 and 10. Here, we report the results obtained using at most 8 burst frames.}. Our burst model gets better perceptual score across all datasets as compared to the ensemble approach, and obtains greater scores for both pixelwise and perceptual scores.

\begin{table}[!t]
\caption{A quantitative comparison of the proposed burst model for varying number of burst images.}
\begin{center}
\begin{tabular}{l@{$\quad$}c@{$\quad$}c@{$\quad$}c@{$\quad$}c@{$\quad$}c}
\toprule
Method & PSNR$\uparrow$ & SSIM$\uparrow$ & LPIPS$\downarrow$ & PieAPP$\downarrow$ & PI$\downarrow$ \\
\midrule
Ours \textit{(S)} & 29.780 & 0.886 & 0.318 & 1.088 & 4.581\\
Ours \textit{(B)} (4 frames) & 29.975 & 0.888 & 0.312 & 1.108 & \textbf{4.437}\\
Ours \textit{(B)} (8 frames) & \textbf{30.043} & \textbf{0.889} & \textbf{0.308} & \textbf{1.076} & 4.461\\
\bottomrule
\end{tabular}
\end{center}
\label{table:ablation_numframes}
\end{table}

\begin{table}[!t]
\caption{A quantitative comparison of the proposed burst model with the ensemble of the single image model.}
\begin{center}
\begin{tabular}{c@{$\;\,$}c@{$\;\,$}c@{$\;\,$}c@{$\;\,$}c@{$\;\,$}c@{$\;\,$}c}
\toprule
Dataset & Method & PSNR$\uparrow$ & SSIM$\uparrow$ & LPIPS$\downarrow$\\
\midrule
\multirow{2}{*}{\rotatebox[origin=c]{90}{Sony}}
& Ours \textit{(E)} (8 frames) & \textbf{30.228} & \textbf{0.893} & 0.323\\
& Ours \textit{(B)} (8 frames) & 30.043 & 0.889 & \textbf{0.308}\\
\midrule
\multirow{2}{*}{\rotatebox[origin=c]{90}{Fuji}}
& Ours \textit{(E)} (8 frames) &  \textbf{28.707} & \textbf{0.853} & 0.522\\
& Ours \textit{(B)} (8 frames) & 28.655 & \textbf{0.853} & \textbf{0.499}\\
\midrule
\multirow{2}{*}{\rotatebox[origin=c]{90}{DRV}}
& Ours \textit{(E)} (8 frames) & 29.074 & 0.917 & 0.353\\
& Ours \textit{(B)} (8 frames) & \textbf{29.104} & \textbf{0.918} & \textbf{0.329}\\
\bottomrule
\end{tabular}
\end{center}
\label{table:ablation_ensemble}
\end{table}

\subsection{Limitations}
\begin{figure}[!t]
    \centering
\subfloat[Traditional pipeline]{
\includegraphics[width=0.475\linewidth]{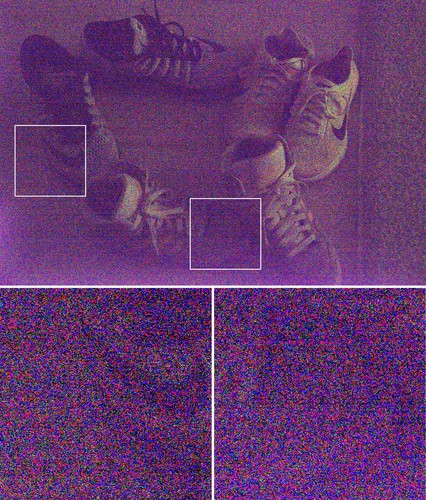}}
\subfloat[Ours (burst)]{
\includegraphics[width=0.475\linewidth]{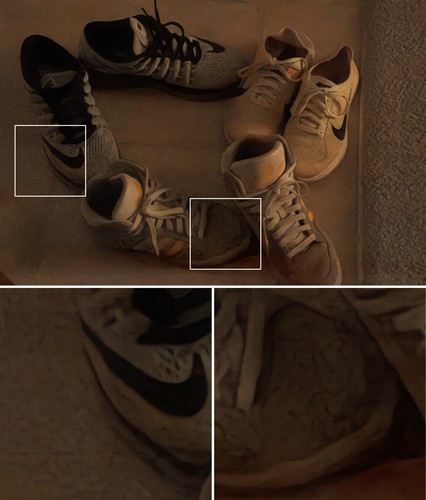}}
\caption{Limitation of our approach. It may sometimes hallucinate false high-frequency details for extremely noisy regions.}
\label{fig:hallucination}
\end{figure}

\begin{figure}[!t]
\centering
\captionsetup[subfigure]{labelformat=empty,farskip=1pt,captionskip=1pt}

\setcounter{subfigure}{0}
\captionsetup[subfigure]{labelformat=empty}

\subfloat[Ours]{
\includegraphics[width=0.475\linewidth]{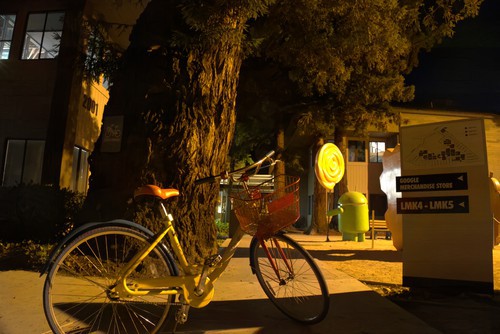}}
\subfloat[Ours + Post-process]{
\includegraphics[width=0.475\linewidth]{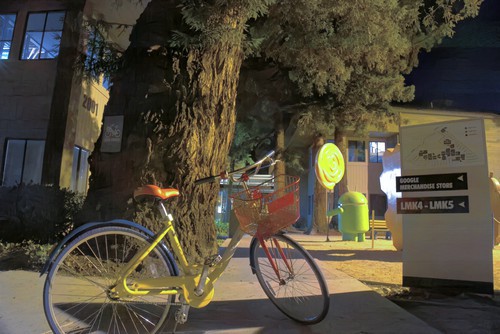}}
\caption{Effect of the post-processing procedure applied to the result of our model for a low-light image captured with 0.1 sec exposure. Post-processing further improves the perceived quality of the enhanced image.}
\label{fig:postprocessing}
\end{figure}

Our approach has a few limitations. First, as illustrated in Fig.~\ref{fig:hallucination}, our model may sometimes hallucinate non-existing high-frequency details. We suspect that this is caused by the excessive noise in the raw images and may be alleviated to some extent by better modeling of the sensor noise. Second, our framework does not explicitly learn to perform white balance correction and tone mapping, and this somewhat affects the results. In an attempt to address this, we employ an additional post-processing step. In particular, we first apply the white balance correction method proposed in~\cite{afifi2019color} to our result. Then, we adjust highlights and shadows using the Core Image API by Apple. Finally, we merge this image with the white-balanced image by using the exposure fusion method by Mertens et al.~\cite{mertens2009exposure} to obtain a tone-mapped image. Fig.~\ref{fig:postprocessing} presents the result of this post-processing step on a sample dark input image. It is evident that this post-processing strategy leads to a visually more pleasing image with vivid colors, further improving the perceived quality of the enhanced image.

\section{Conclusion}
In this study, we tackle the problem of learning to generate long-exposure images from a set of extremely low-light burst images. We developed a new deep model that incorporates a coarse-to-fine strategy to better enhance the details of the output. Moreover, we extended this network architecture to work with a burst of images via a novel a permutation invariant CNN architecture, which efficiently processes the information exchanged between the features of the burst frames. Our experiments show that our burst method achieves higher quality results than the state-of-the-art models, better capturing finer details, texture and color information and reducing noise. That being said, our analysis also suggests that there is still much room for improvement, especially for ultra-low light scenes. 

\section*{Acknowledgments}
This work was supported in part by GEBIP 2018 Award of the Turkish Academy of Sciences to E. Erdem, BAGEP 2021 Award of the Science Academy to A. Erdem. We would like to thank KUIS AI Center for letting us use their High Performance Computing Cluster.

\ifCLASSOPTIONcaptionsoff
  \newpage
\fi

\bibliographystyle{IEEEtran} 
\bibliography{tip}

\vskip -2\baselineskip plus -1fil

\begin{IEEEbiography}[{\includegraphics[width=1in]{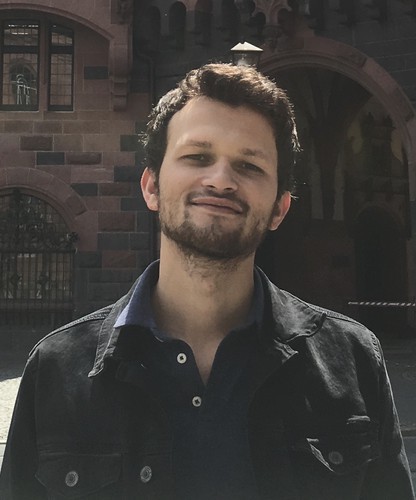}}]{Ahmet Serdar Karadeniz}
received the B.Sc. degree in Mathematics from Middle East Technical University, Ankara, Turkey, in 2018. He is currently an M.Sc. student in the Department of Computer Engineering at Hacettepe University, Ankara, Turkey. His research interests include machine learning, image processing and computational photography.
\end{IEEEbiography}

\vskip -2\baselineskip plus -1fil
\begin{IEEEbiography}[{\includegraphics[width=1in,clip,keepaspectratio]{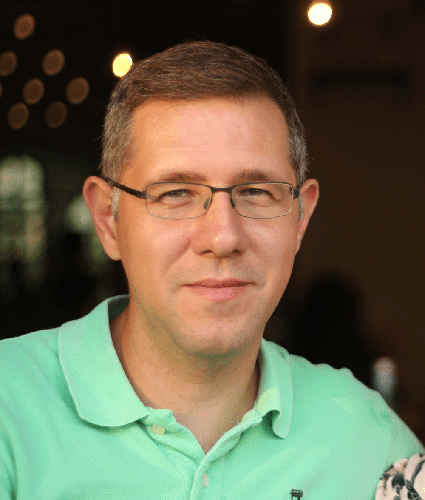}}]{Erkut Erdem} received his Ph.D. degree from Middle East Technical University in 2008. After completing his Ph.D., he continued his post-doctoral studies with Télécom ParisTech, École Nationale Supérieure des Télécommunications, France, from 2009 to 2010. He has been an Associate Professor with the Department of Computer Engineering, Hacettepe University, Turkey, since 2014. His research interests include semantic image editing, visual saliency prediction, and integrated vision and language applications.
\end{IEEEbiography}

\vskip -2\baselineskip plus -1fil
\begin{IEEEbiography}[{\includegraphics[width=1in,clip,keepaspectratio]{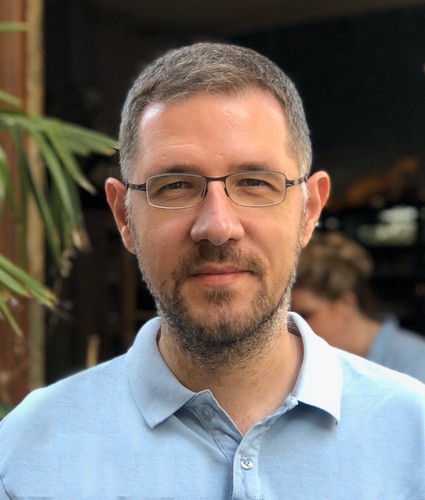}}]{Aykut Erdem} is an Associate Professor of Computer Science at Koç University. He received his Ph.D. degree from Middle East Technical University in 2008. He was a post-doctoral researcher at the Ca’Foscari University in Venice in the EU-FP7 SIMBAD project, from 2008 to 2010. Previously, he was with the Computer Engineering Department at Hacettepe University. The broad goal of his research is to explore better ways to understand, interpret and manipulate visual data. His current research focuses on investigating learning-based approaches to image editing, visual saliency estimation, and connecting vision and language.
\end{IEEEbiography}

\end{document}